\documentclass[10pt,twocolumn,letterpaper]{article}

\usepackage{iccv}
\usepackage{times}
\usepackage{epsfig}
\usepackage{graphicx}
\usepackage{amsmath}
\usepackage{amssymb}
\usepackage{booktabs}

\usepackage[dvipsnames]{xcolor}
\usepackage{colortbl}
\definecolor{gray}{gray}{0.5}

\usepackage[utf8]{inputenc}

\usepackage{atbegshi,etoolbox}
\makeatletter
\newcommand{\discardpages}[1]{
	\xdef\discard@pages{#1}
	\AtBeginShipout{
		\renewcommand*{\do}[1]{
			\ifnum\value{page}=##1\relax%
			\AtBeginShipoutDiscard
			\gdef\do####1{}
			\fi%
		}%
		\expandafter\docsvlist\expandafter{\discard@pages}
	}%
}
\newif\ifkeeppage
\newcommand{\keeppages}[1]{
	\xdef\keep@pages{#1}
	\AtBeginShipout{
		\keeppagefalse%
		\renewcommand*{\do}[1]{
			\ifnum\value{page}=##1\relax%
			\keeppagetrue
			\gdef\do####1{}
			\fi%
		}%
		\expandafter\docsvlist\expandafter{\keep@pages}
		\ifkeeppage\else\AtBeginShipoutDiscard\fi
	}%
}
\makeatother

\DeclareMathOperator*{\argmax}{arg\,max}

\DeclareMathOperator*{\median}{median}
\DeclareMathOperator*{\softmax}{softmax}

\usepackage{umoline}

\usepackage{enumitem}

\graphicspath{{./}{graphics/}}

\usepackage[pagebackref=true,breaklinks=true,letterpaper=true,colorlinks,bookmarks=false]{hyperref}

\iccvfinalcopy 

\def\iccvPaperID{9923} 

\ificcvfinal\fi

\begin{document}

\title{Identification of Systematic Errors of Image Classifiers on Rare Subgroups}

\author{Jan Hendrik Metzen$^{(1)}$, Robin Hutmacher$^{(1)}$, N.\,Grace Hua$^{(1)}$, Valentyn Boreiko$^{(1, 2)}$, Dan Zhang$^{(1)}$\\
(1) Bosch Center for Artificial Intelligence, Robert Bosch GmbH (2) University of T{\"u}bingen \\
{\tt\footnotesize \{janhendrik.metzen,robin.hutmacher,grace.hua,dan.zhang2\}@de.bosch.com;
 valentyn.boreiko@uni-tuebingen.de
}
}

\maketitle
\ificcvfinal\thispagestyle{empty}\fi

\begin{abstract}
	Despite excellent average-case performance of many image classifiers, their performance can substantially deteriorate on semantically coherent subgroups of the data that were under-represented in the training data. These \emph{systematic errors} can impact both fairness for demographic minority groups as well as robustness and safety under domain shift.  A major challenge is to identify such subgroups with subpar performance when the subgroups are not annotated and their occurrence is very rare. We leverage recent advances in \emph{text-to-image models} and search in the space of textual descriptions of subgroups (``prompts'') for subgroups where the target model has low performance on the prompt-conditioned synthesized data. To tackle the exponentially growing number of subgroups, we employ \emph{combinatorial testing}. We denote this procedure as \textsc{PromptAttack} as it can be interpreted as an adversarial attack in a prompt space. We study subgroup coverage and identifiability with \textsc{PromptAttack} in a controlled setting and find that it identifies  systematic errors with high accuracy. Thereupon, we apply \textsc{PromptAttack} to ImageNet classifiers and identify novel systematic errors on rare subgroups. 
\end{abstract}

\section{Introduction}
\label{sec:intro}

\begin{figure}[tb]
	\begin{center}
		\includegraphics{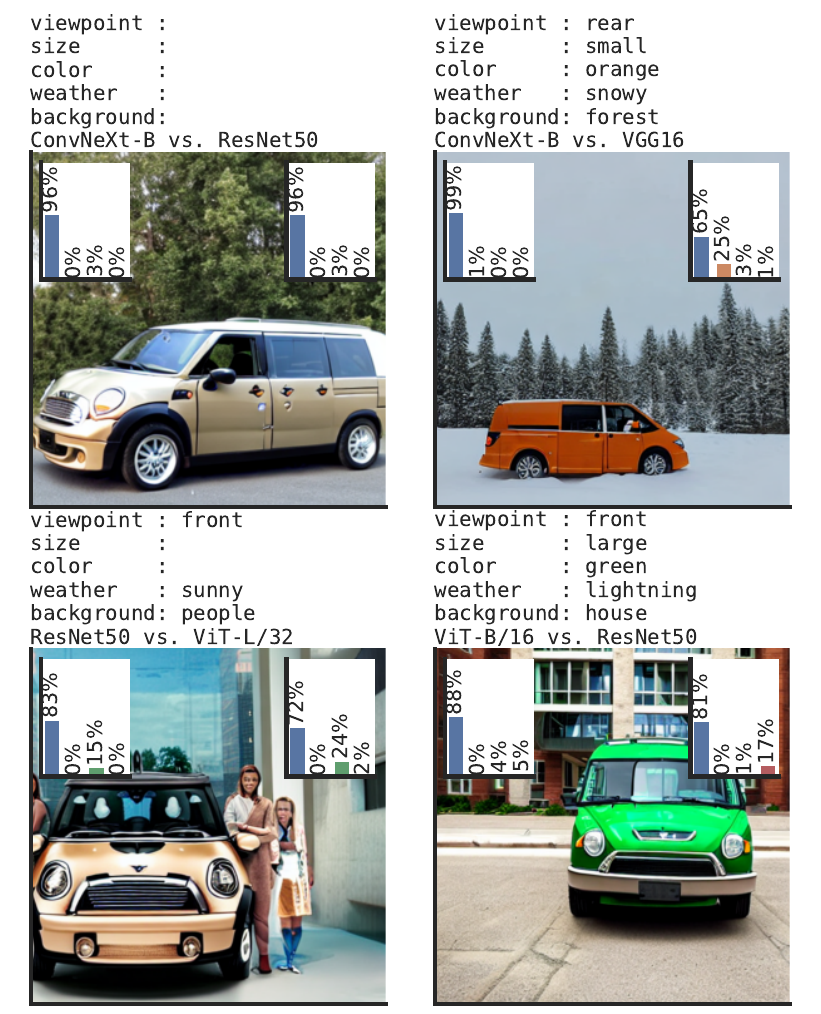}
		\caption{Samples along with histograms over two models' class prediction rates (shown in left and right inlays, based on 400 samples) for 4 different subgroups. The baseline subgroup (top left) is classified mostly as \textcolor{blue}{minivan} by all models, while the misclassification rates to a \textcolor{RedOrange}{snowplow} (top right), to \textcolor{OliveGreen}{pickup} (bottom left), and to \textcolor{BrickRed}{police\_van} (bottom right) are significantly increased on the shown subgroups for a VGG16, ViT-L/32, ResNet50 respectively.  We refer to Section \ref{sec:experiments_imagenet} and \ref{sec:samples} for more details and samples.}
		\label{Figure:syserror_minivan}
	\end{center}
    \vspace*{-.5cm}
\end{figure}

Deep learning based approaches have revolutionized many fields of computer vision \cite{liu_deep_2020,9356353} and are increasingly applied in safety-critical applications such as automated driving \cite{grigorescu2020autonomous_driving}. An important prerequisite for deployment of learned models in such safety-critical domains is that they need to work reasonably well for all subgroups from an operational design domain \cite{operational_design_domain} and strong requirements are imposed on assuring safety of systems \cite{Blank_Hueger_Mock_Stauner_2022}. 
That is: there must not be catastrophic but avoidable failure cases on any subgroup, regardless of how rare the subgroup might be.
Unfortunately, spurious correlations in the training data can often result in classifiers that utilize shortcut decision making \cite{geirhos2020shortcut,unbiased_look_at_dataset_bias,li_2021_discover,Li_2022_discover} - a phenomenon long known from animal and human psychology \cite{cleverhans}. Such shortcuts can work well on subgroups that occur frequently  in-distribution, that is: on data that follows the same distribution as the training data. However, they can easily fail after a domain shift to out-of-distribution data since very rare subgroups become suddenly much more frequent \cite{pmlr-v139-zhou21g}. For instance, Beery et al.\,\cite{Beery_2018_ECCV} demonstrate that a shift in background can largely affect an image classifier, resulting in misclassifying, e.g., a cow on the beach.

Accordingly, a crucial aspect of \emph{model auditing} \cite{model_auditing} is to separately evaluate the behaviour of a classifier on every subgroup from a large set of subgroups. If the performance of a classifier on certain subgroups is considerably worse than on the totality of the domain's data, then we denote this subgroup as a \emph{systematic error} of a classifier \cite{eyuboglu_domino:_2022,jain_distilling_2022,wiles2022discovering}. More specifically, a systematic error refers to a subgroup of inputs on which a pretrained classifier has a high probability of misclassification (``error'') and at the same time a large percentage of elements in the subgroup share a human-interpretable concept: the group appears semantically coherent to a human (``systematic'').  Applying methods for identifying such systematic errors could become a prerequisite for deployment in many domains while at the same time, systematic error are \emph{actionable}: their exemplars can be used for finetuning a model and improving its robustness, reliability, and fairness \cite{gao2022adaptive}.

Some prior work \cite{eyuboglu_domino:_2022,jain_distilling_2022} require the availability of a labelled hold-out set covering data from rare subgroups for identification of systematic errors on these subgroups.  Unfortunately, it is often expensive to acquire (labelled) data for certain subgroups that are very rare in the domain pre-deployment, even though subgroups could become much more frequent after some domain shift. 
This is problematic because systematic errors are much more likely to occur on subgroups that are rare in the training data. Other prior work is based on large unlabelled hold-out data but requires a human-in-the-loop \cite{gao2022adaptive}, which increases the costs of systematic error identification and thus limits applicability.


Another line of work (including ours) focuses on auditing models on synthetically generated subgroup data. Recent progress of \emph{text-to-image models} \cite{dalle2,rombach2021highresolution,saharia2022photorealistic,muse} in terms of compositionality can allow synthesizing data of rare subgroups that have not been part of the training data. Wiles et al.\,\cite{wiles2022discovering} focused on an open-ended approach that synthesizes data according to the distribution induced by a fixed prompt that encodes the class but no subgroup information. Concurrently to our work, Vendrow et al.\,\cite{vendrow2023dataset} generated text-conditioned counterfactual examples to study the robustness to single semantic shifts.

\begin{figure*}
	\begin{center}
		\includegraphics[width=.9\linewidth]{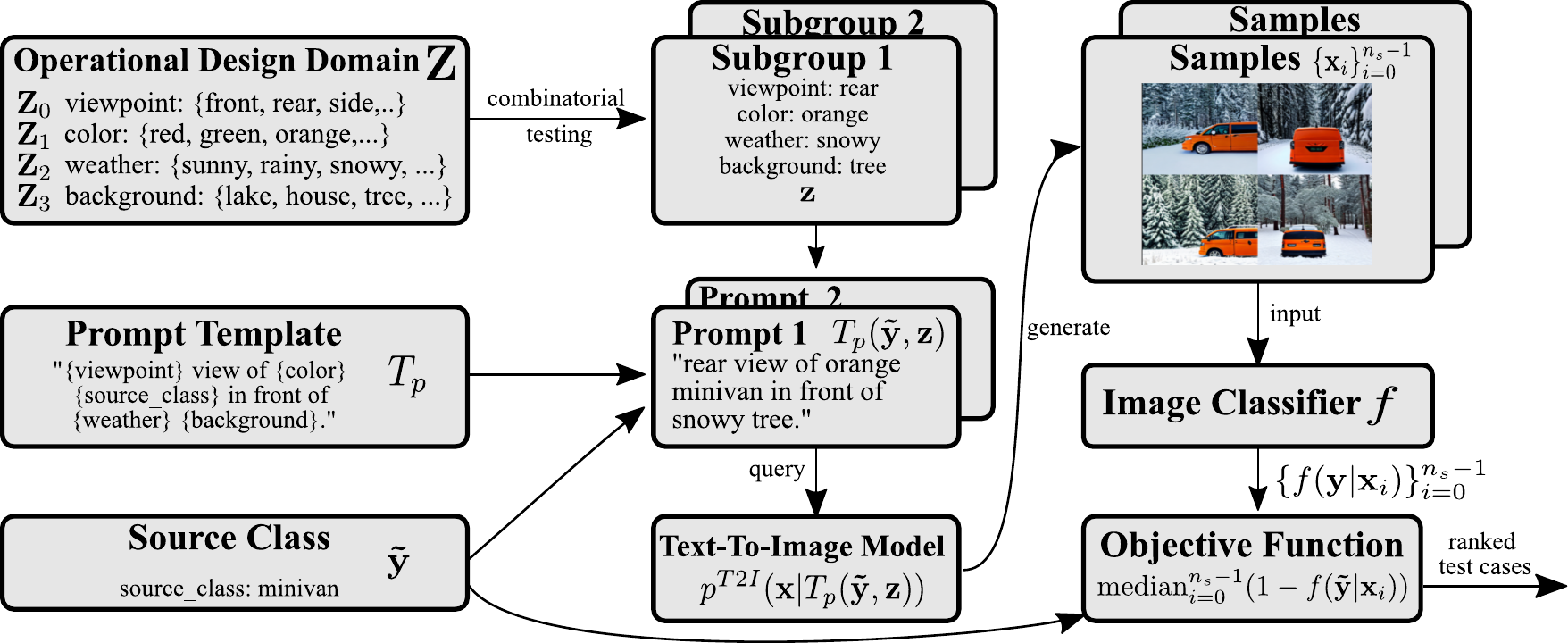}
	\end{center}
    \vspace*{-.25cm}
	\caption{Illustration of \textsc{PromptAttack}: domain experts define an operational design domain $\mathbf{Z}$ consisting of semantic dimensions $\mathbf{Z}_j$. Combinatorial testing is used to generate a set of subgroups. A prompt $T_p(\mathbf{\tilde{y}},\mathbf{z})$ is instantiated from a prompt template $T_p$ based on the respective subgroup $\mathbf{z}$ and source class $\mathbf{\tilde{y}}$. A text-to-image model $p^{T2I}$ generates $n_s$ samples $\{\mathbf{x}_i\}_{i=0}^{n_s-1}$ for the prompt. The image classifier $f$ under investigation predicts class probabilities $\{f(\mathbf{y} \vert \mathbf{x}_i)\}_{i=0}^{n_s-1}$ for the samples. An objective function provides a ranking of subgroups based upon the source class predictions $f(\mathbf{\tilde{y}} \vert \mathbf{x}_i)$, where low median class score for the source class indicates a potential systematic error.}
	\label{graphics:prompt_attack_illustration}
	\vspace*{-.25cm}
\end{figure*}

We propose \textsc{PromptAttack} (see Figure \ref{graphics:prompt_attack_illustration}), which leverages text-to-image models for synthesizing images of subgroups by encoding subgroup information directly in the prompt. To deal with large operational design domains and the resulting combinatorial explosion of subgroups, \textsc{PromptAttack} builds upon \emph{combinatorial testing} \cite{Nie_Leung_ACM,8102999} that allows a near-equable coverage of the operational design domain while keeping the number of explored subgroups relatively small. 
In contrast to the open-ended approach by Wiles et al.\,\cite{wiles2022discovering}, \textsc{PromptAttack} is targeted and reliably explores subgroups from a prespecified operational design domain (see Section \ref{sec:coverage}). Moreover, it does not require any pretrained models or heuristics components for failure case clustering and captioning. In contrast to Vendrow et al.\,\cite{vendrow2023dataset}, \textsc{PromptAttack} can identify systematic errors on subgroups that require the concurrence of several semantic shifts (see Figure \ref{Figure:teaser}).



Overall, our main contributions are the following:
\begin{itemize}[noitemsep,topsep=0pt,parsep=0pt,partopsep=0pt]
	\item In Section \ref{sec:method}, we introduce \textsc{PromptAttack}, a novel procedure for identifying systematic errors based on synthetic data from a text-to-image model, conditioned on a prompt encoding subgroup and class information. \textsc{PromptAttack} explores a large subset of subgroups from an operational design domain using combinatorial testing, achieving near-equable coverage of subgroups (Section \ref{sec:coverage}).
	\item We propose a benchmark for testing and comparing methods for systematic error identification (Section \ref{sec:benchmark}). In contrast to prior work \cite{eyuboglu_domino:_2022}, this benchmark does not train multiple classifiers with training-time interventions but is based purely on inference-time interventions on zero-shot classifiers such as CLIP \cite{pmlr-v139-radford21a}.
	\item \textsc{PromptAttack} identifies classifier-specific and targeted systematic misclassifications on rare subgroups of ImageNet classifiers (see Figure \ref{Figure:syserror_minivan} and Section \ref{sec:experiments_imagenet}).
\end{itemize}

\section{Related Work}
\label{sec:relatedwork}

We review related work in the computer vision domain while noting that identification of systematic errors and harmful behaviour is an important topic in other fields such as large language models as well (``red teaming'') \cite{perez2022redteaming,ganguli2022red}.

\textbf{Building upon Subgroup Annotation.}
Several prior works have investigated performance on datasets where information on certain semantic dimensions is available for each datapoint and thus direct evaluation of subgroup error is feasible. 
For instance, Hendrycks et al.\,\cite{hendrycks2021manyfaces} collected four real-world datasets containing semantic dimensions like \emph{artistic renditions} (ImageNet-R), \emph{country}, \emph{year}, and \emph{camera} (StreetView StoreFronts), or \emph{object size, object occlusion, camera viewpoint, and camera zoom} (DeepFashion Remixed). The influence of \emph{image background} can be studied based on ImageNet-9 \cite{xiao2021noise} and Waterbirds \cite{Sagawa2020Distributionally}. WildDash \cite{Zendel_2018_ECCV} allows studying the impact of different \emph{visual hazards}. ImageNet-X \cite{Idrissi2022ImageNetX} adds sixteen human annotations
of semantic dimensions such as \emph{pose}, \emph{background}, or \emph{lighting} to each ImageNet-1k validation sample. %
Such approaches require large efforts in data collection and subgroup annotation and have thus limited scalability and flexibility. Moreover, if there are several interacting semantic dimensions, then the number of datapoints required to achieve a full coverage of the design domain grows exponentially. One partial remedy to this combinatorial explosion is combinatorial t-wise testing  \cite{Nie_Leung_ACM,8102999,Gladisch_2020_CVPR_Workshops}. Alternatively, synthetic corruptions can be applied to existing images resulting in a semantic \emph{corruption} dimension (ImageNet-C) \cite{hendrycks2018benchmarking}. 
However, not all semantic dimensions can be simulated. 



\textbf{Failure Identification without Subgroup Annotation.}
Since explicit annotation of semantic dimension is costly, recent works have focused on automating the process of identifying systematic errors. One line of work resorts to a labeled hold-out set. Coherent groups of errors on the hold-out set can be identified by error-aware soft-clustering on the final feature space of the classifier \cite{dEon2022Spotlight}. However, this does not provide an interpretation of the subgroup. Leveraging the text-image embedding alignment in CLIP \cite{pmlr-v139-radford21a}, both \cite{eyuboglu_domino:_2022} and \cite{jain_distilling_2022} operated in the latent space of CLIP to identify semantically coherent subgroups and generate human interpretable subgroup annotation. The main disadvantage of these works is that they require the availability of a labeled hold-out set. Since systematic errors are more likely to occur on untypical/rare data \cite{eyuboglu_domino:_2022}, identifying them requires a hold-out dataset containing such cases, which is unrealistic.
AdaVision \cite{gao2022adaptive} introduced a human-in-the-loop process to discover systematic errors by adaptively querying real images from LAION-5B \cite{schuhmann2022laionb} (via CLIP similarity). In contrast to AdaVision, our procedure does not require a human-in-the-loop, which can be preferred if model auditing needs to be done regularly or for a large number of models.





The most similar line of work to ours is built on top of text-to-image synthesis models \cite{dalle2,rombach2021highresolution,saharia2022photorealistic,muse}. For instance, in \cite{Kattakinda2022invariant,lynch2022evaluating,vendrow2023dataset}, counterfactual examples are generated according to the input text, which indicates the semantic shift, e.g., background, lighting, or style. These works considered a single semantic shift, whereas systematic errors can result from compounding shifts in multiple semantic factors. Wiles et al.\,\cite{wiles2022discovering} did not pre-specify the semantic shift but iteratively synthesize samples based on a text description, cluster the failure cases, and refine the text description. 
In contrast to such an ``open-ended'' search, our work focuses on finding failures within an operational design domain and achieves high coverage of that domain (see Section \ref{sec:coverage}). Moreover, our approach is conceptually more simple and does not require clustering and captioning of failure cases. Both approaches can be seen as complementary.

\section{Method}
\label{sec:method}
In this section, we introduce our proposed procedure \textsc{PromptAttack}; see Figure \ref{graphics:prompt_attack_illustration} for an illustration.

\subsection{Background}
We consider image classifiers $f: \mathbb{X} \times \mathbb{Y} \rightarrow [0, 1]$ and denote the predicted probability of class $\mathbf{y} \in \mathbb{Y} = \{1, \dots, C\}$ for image $\mathbf{x} \in \mathbb{X}$ by $f(\mathbf{y} \vert \mathbf{x})$. We assume that we operate in a domain where $\mathbf{x}, \mathbf{y}$ are governed by a distribution $p(\mathbf{x}, \mathbf{y})$. We are interested in exploring properties of $f$ on semantically coherent subgroups of the data manifold, which we formalize by conditioning on some latent $\mathbf{z}$: $p(\mathbf{x}, \mathbf{y} \vert \mathbf{z})$. For brevity, we also denote the subgroups themselves by $\mathbf{z}$. 
We note that in contrast to Wiles et al.\,\cite{wiles2022discovering}, we do not build on the conditional distribution $p(\mathbf{z} \vert \mathbf{x}, \mathbf{y})$ and thus do not require an image captioning model.

\subsection{Systematic Errors: Definition}
We define the risk of a classifier $f$ on a subgroup $\mathbf{z}$ by $R_f(\mathbf{z}) = \mathbb{E}_{p(\mathbf{x}, \mathbf{y} \vert \mathbf{z})} L(f(\cdot \vert \mathbf{x}), \mathbf{y}) $, where $\mathbb{E}_{p}$ denotes the expectation over $p$ and $L: [0, 1]^C \times \mathbb{Y} \mapsto \mathbb{R}$ is a loss function. Moreover, we set the baseline (irreducible) risk on $\mathbf{z}$ to $R_\mathcal{B}(\mathbf{z}) = \mathbb{E}_{p(\mathbf{x}, \mathbf{y} \vert \mathbf{z})} L(p(\cdot \vert \mathbf{x}, \mathbf{z}), \mathbf{y}) $, with $p(\mathbf{y} \vert \mathbf{x}, \mathbf{z}) = p(\mathbf{x}, \mathbf{y} \vert \mathbf{z}) / p(\mathbf{x} \vert \mathbf{z})$. We note $R_f(\mathbf{z}) \geq R_\mathcal{B}(\mathbf{z})$.
We assume that we are provided with a predefined set of subgroups $\mathbf{Z}$ that we denote as the \emph{operational design domain} \cite{operational_design_domain}. We are interested in subgroups $\mathbf{z} \in \mathbf{Z}$ on which a classifier $f$ has high risk $R_f(\mathbf{z})$ while the baseline risk $R_\mathcal{B}(\mathbf{z})$ remains low. If the subgroups $\mathbf{z}$ are designed in a way to encourage semantic coherence, such high risk subgroups are human interpretable and actionable. 
More specifically, we rank subgroups $\mathbf{z} \in \mathbf{Z}$ based on $R(\mathbf{z}) = R_f(\mathbf{z}) - R_\mathcal{B}(\mathbf{z})$; top-ranked $\mathbf{z}$ with sufficiently high risk are \emph{systematic errors}. Similarly, we define a \emph{systematic misclassification} into class $\mathbf{y}^{(t)}$ by $R_f(\mathbf{z}, \mathbf{y}^{(t)}) = \mathbb{E}_{p(\mathbf{x}, \mathbf{y} \vert \mathbf{z})} L(f(\cdot \vert \mathbf{x}), \mathbf{y}^{(t)}) 1_{[\mathbf{y} \neq \mathbf{y}^{(t)}]}$ and $R_\mathcal{B}(\mathbf{z}, \mathbf{y}^{(t)})$ analogously, with $1_{[\mathbf{y} \neq \mathbf{y}^{(t)}]}$ being the indicator function of $\mathbf{y} \neq \mathbf{y}^{(t)}$. The top-ranked $\mathbf{z}$ according to $R(\mathbf{z}, \mathbf{y}^{(t)}) = R_f(\mathbf{z}, \mathbf{y}^{(t)}) - R_\mathcal{B}(\mathbf{z}, \mathbf{y}^{(t)})$ are systematic misclassifications into $\mathbf{y}^{(t)}$ for sufficiently high $R(\mathbf{z}, \mathbf{y}^{(t)})$. We note the risk $R$ can be made invariant to the classifier's calibration (e.g., for $L$ being a 0-1 loss function) or sensitive to it (for most other choices of $L$).

\subsection{Systematic Errors: Approximations} \label{Subsection:Approximations}
We make several approximations and assumptions for tractable systematic error identification; empirical evidence in Section \ref{sec:quantitative_evaluation} suggests these hold reasonably well in practice.

\textbf{1. Monte Carlo Approximation.} 
In general we cannot compute $\mathbb{E}_{p(\mathbf{x}, \mathbf{y} \vert \mathbf{z})}$ in the definition of $R_f(\mathbf{z})$. Thus,  we resort to approximating the expectation based on $n_s$ samples $\mathbf{x}_i, \mathbf{y}_i \sim p(\mathbf{x}, \mathbf{y} \vert \mathbf{z})$: $R_f(\mathbf{z}) \approx \frac{1}{n_s} \sum\limits_{i=0}^{n_s-1} L(f(\cdot \vert \mathbf{x}_i), \mathbf{y}_i).$

\textbf{2. Synthetic Data.}
Real-world samples $\mathbf{x}, \mathbf{y} \sim p(\mathbf{x}, \mathbf{y} \vert \mathbf{z})$ from semantically coherent subgroups $\mathbf{z}$ are typically not available for two reasons: (i) typical real-world data $\mathbf{x}, \mathbf{y}$ lacks human annotated information on $\mathbf{z}$, such as captions in a given format or subgroup annotations. (ii) Even if $\mathbf{z}$ were available or could be inferred, the coverage of the operational design domain $\mathbf{Z}$ can be very low: some rare subgroups $\mathbf{z} \in \mathbf{Z}$ will have low $p(\mathbf{z} \vert \mathbf{x}, \mathbf{y})$ and may not be represented at all in finite sample sets from $p(\mathbf{x}, \mathbf{y})$.
However, the performance of $f$ on such rare subgroups can still be highly relevant in safety-critical applications, as specifically rare corner cases may be the ones where the generalization of a classifier $f$ fails. Because of this, we resort to sampling $\mathbf{x}, \mathbf{y}\vert \mathbf{z}$ from learned approximations $\hat{p}$ of the real-world data distribution. For this, we leverage recent progress on text-to-image generative model  $p^{T2I}(\mathbf{x} \vert t)$ such as Stable Diffusion \cite{rombach2021highresolution}, which condition image generation on a text prompt $t$, as detailed below. 

\textbf{3. Sampling Class-Conditional.}
We can use text-to-image models for sampling from $\hat{p}(\mathbf{x} \vert \mathbf{z})$ by representing the subgroup $\mathbf{z}$ as a text prompt $t$. However, sampling from $\hat{p}(\mathbf{x}, \mathbf{y} \vert \mathbf{z}) = \hat{p}(\mathbf{y} \vert \mathbf{x}, \mathbf{z}) \hat{p}(\mathbf{x} \vert \mathbf{z})$ would also require an approximation $\hat{p}(\mathbf{y} \vert \mathbf{x}, \mathbf{z})$, that is: the conditional probability of a specific class $\mathbf{y}$ given an image $\mathbf{x}$ and subgroup $\mathbf{z}$. Such an approximation $\hat{p}(\mathbf{y} \vert \mathbf{x}, \mathbf{z})$ is not generally available or easily estimated (estimating $\hat{p}(\mathbf{y} \vert \mathbf{x}, \mathbf{z})$ from data would require a large number of samples $(\mathbf{x}, \mathbf{y})$ annotated with $\mathbf{z}$, which we precluded above).  
Instead, we explicitly condition the generation of $\mathbf{x}$ on a desired class $\mathbf{\tilde{y}}$. Effectively, this corresponds to focusing on systematic errors on a specific source class $\mathbf{\tilde{y}}$. We realize the approximate $\hat{p}(\mathbf{x} \vert \mathbf{\tilde{y}}, \mathbf{z})$ by including class information $\mathbf{\tilde{y}}$  along with $\mathbf{z}$ in a text prompt $t$  as detailed in Section \ref{section:operational_design_domain}.

\textbf{4. Negligible Baseline Risk.}
We cannot evaluate $R_\mathcal{B}(\mathbf{z})$ directly since $p(\mathbf{y} \vert \mathbf{x}, \mathbf{z})$ is unavailable. Because of this, we limit ourselves to choices of $\mathbf{Z}$ where by design for every $\mathbf{z} \in \mathbf{Z}$, we have for $\mathbf{x} \sim p(\mathbf{x} \vert \mathbf{\tilde{y}}, \mathbf{z})$ that $p(\mathbf{y} \vert \mathbf{x}, \mathbf{z}) \approx 1 \text{ if } \mathbf{y} = \mathbf{\tilde{y}} \text{ else } 0$. That is: classes do not overlap on $\mathbf{z}$ and images $\mathbf{x}\vert \mathbf{\tilde{y}}, \mathbf{z}$ belong unambiguously to the same class $\mathbf{\tilde{y}}$. Accordingly, the baseline risk $R_\mathcal{B}(\mathbf{z})$ on $p(\mathbf{x}, \mathbf{y} \vert \mathbf{z})$ is negligibly small for typical loss functions (unlike on the unconditional $p(\mathbf{x}, \mathbf{y})$) and we can approximate $R(\mathbf{z}) \approx R_f(\mathbf{z})$ and $R(\mathbf{z}, \mathbf{y}^{(t)}) \approx R_f(\mathbf{z}, \mathbf{y}^{(t)})$.

\textbf{Dealing with Violations.}
We note that the above approximations do not hold strictly as the generative model $\hat{p}(\mathbf{x} \vert \mathbf{\tilde{y}}, \mathbf{z})$ will not perfectly approximate the real-data subgroups $p(\mathbf{x} \vert \mathbf{\tilde{y}}, \mathbf{z})$: it  may generate (i) valid data from $p(\mathbf{x} \vert \mathbf{\tilde{y}})$ that is ``out-of-subgroup'' (OOS), that is: has very low probability under $p(\mathbf{x} \vert \mathbf{\tilde{y}}, \mathbf{z})$, (ii) data that does not belong to the class $\mathbf{\tilde{y}}$, that is: low $p(\mathbf{x} \vert \mathbf{\tilde{y}})$ (``out-of-class'', OOC), and (iii) data that is even very unlikely under $p(\mathbf{x})$ (OOD sampling \cite{wiles2022discovering}). Recent progress in text-to-image models, for which $\hat{p}(\mathbf{x} \vert \mathbf{\tilde{y}}, \mathbf{z})$ more closely approximates $p(\mathbf{x} \vert \mathbf{\tilde{y}}, \mathbf{z})$, makes such OOS/OOC/OOD samples occur less often. To reduce them further, we carefully engineer text prompts for $\mathbf{\tilde{y}}$ and $\mathbf{z}$ (see Section \ref{section:operational_design_domain}). This requirement for ``prompt engineering'' is a shortcoming but we are optimistic that future text-to-image models will reduce its need.

Even with careful prompt engineering, a few OOS/OOC/OOD samples might still dominate the Monte-Carlo estimate for $R_f(\mathbf{z})$. We thus resort to robust estimators of central tendency for $R_f(\mathbf{z})$, which are less affected by outliers, such as $R_f(\mathbf{z}) \approx \median_{i=0}^{n_s-1} L(f(\cdot \vert \mathbf{x}_i), \mathbf{\tilde{y}})$. 

\subsection{Operational Design Domain $\mathbf{Z}$} \label{section:operational_design_domain}
In principle, our approach allows arbitrary operational design domains $\mathbf{Z}$. We specifically focus on a setting where the operational design domain is compositional: $\mathbf{Z}=\mathbf{Z}_0 \times \dots \times \mathbf{Z}_{n_Z - 1}$, where every $\mathbf{Z}_i$ corresponds to a  semantically meaningful dimension. Every $\mathbf{z} \in \mathbf{Z}$ is then a tuple containing $n_Z$ values (one for each semantic dimension). 
As we use text-to-image models $p^{T2I}(\mathbf{x} \vert t)$ for sampling from a subgroup, we assume the operational design domain comes with a \emph{prompt template} $T_p$ that allows mapping this $n_Z$-dimensional tuple along with the class $\mathbf{\tilde{y}}$ to a text prompt: $t = T_p(\mathbf{\tilde{y}}, \mathbf{z})$. We thus set $\hat{p}(\mathbf{x} \vert \mathbf{\tilde{y}}, \mathbf{z}) = p^{T2I}(\mathbf{x} \vert T_p(\mathbf{\tilde{y}}, \mathbf{z}))$. We note that the choice of the prompt template $T_p$ can significantly affect the efficacy of our procedure.

If we specify the operational design domain $\mathbf{Z}$ as above, we have $\vert \mathbf{Z} \vert = \prod_{i=0}^{n_Z - 1} \vert \mathbf{Z}_i \vert$. Accordingly, the number of subgroups in the operational domain grows exponentially with the number of semantic dimensions $n_Z$. To deal with large $n_Z$, we optionally employ \emph{combinatorial testing} \cite{Nie_Leung_ACM,8102999} to test only a subset of subgroups $\mathbf{Z}_C \subseteq \mathbf{Z}$ for systematic errors. Specifically, for a value $n_C \leq n_Z$, combinatorial testing ensures that for any $i_0,\dots, i_{n_C - 1} \leq n_Z - 1$ and for all $\mathbf{z} \in \mathbf{Z}$  there exists a $\mathbf{z}^C \in \mathbf{Z}_C$ such that $\mathbf{z}_{i_0} = \mathbf{z}^C_{i_0}$, ..., $\mathbf{z}_{i_{n_C - 1}} = \mathbf{z}^C_{i_{n_C - 1}}$. That is, for every combination of $n_C$ semantic dimensions $\mathbf{Z}_i$, every possible combination of values from these dimensions is covered at least once in $\mathbf{Z}_C$. Choosing  $n_C < n_Z$ reduces the number of tested subgroups at the cost of not reaching a full (but near-equable) coverage of $\mathbf{Z}$. 
Combinatorial testing allows evaluating different loss functions concurrently; we use $L(f(\cdot \vert \mathbf{x}), \mathbf{\tilde{y}}) = 1 - f(\mathbf{\tilde{y}} \vert \mathbf{x})$ for systematic errors and $L(f(\cdot \vert \mathbf{x}), \mathbf{y}^{(t)}) = f(\mathbf{y}^{(t)} \vert \mathbf{x})$ for systematic misclassifications for multiple choices of $\mathbf{y}^{(t)}$.


\section{Quantitive Evaluation}
\label{sec:quantitative_evaluation}
We perform quantitive evaluations in terms of coverage properties of an operational design domain (Section \ref{sec:coverage}) and the ability of \textsc{PromptAttack} to recover known systematic errors from a zero-shot classifier (Section \ref{sec:benchmark}).

\subsection{Coverage Analysis of Conditional versus Unconditional Synthesis}
\label{sec:coverage}

\textbf{Motivation.} The primary motivation for \textsc{PromptAttack} is to encourage full exploration of an operational design domain $\mathbf{Z}$ by explicitly conditioning image generation on subgroups $\mathbf{z} \in \mathbf{Z}$, that is, to sample from $\hat{p}(\mathbf{x} \vert \mathbf{z})$ rather than unconditionally from $\hat{p}(\mathbf{x})$  as done by Wiles et al.\,\cite{wiles2022discovering} (we skip the conditioning on $\mathbf{\tilde{y}}$ here for brevity). We investigate in this subsection whether this indeed results in better coverage: $\hat{p}(\mathbf{z}) = \int_\mathbf{x} \hat{p}(\mathbf{z} \vert \mathbf{x}) \hat{p}(\mathbf{x}) \, d\mathbf{x}$ should be near uniform over $\mathbf{Z}$ when $\hat{p}(\mathbf{x})$ is the empirical distribution of samples generated by \textsc{PromptAttack}. For this analysis (but not for \textsc{PromptAttack} itself), we need a mechanism to estimate $\hat{p}(\mathbf{z} \vert \mathbf{x})$, for which we employ a zero-shot classifier derived from the multimodal image-text model CLIP \cite{pmlr-v139-radford21a}.

\begin{figure}[tb]
\includegraphics{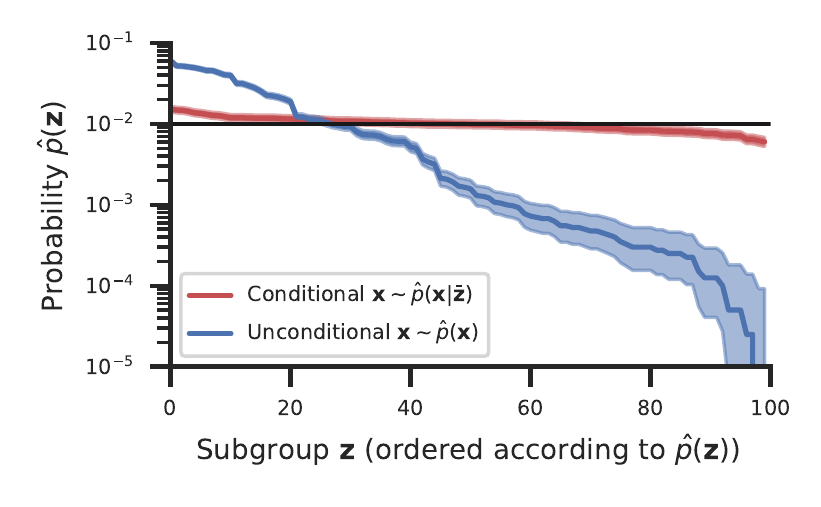}
    \vspace*{-.5cm}
	\caption{Estimate of $\hat{p}(\mathbf{z}) = \int_\mathbf{x} \hat{p}(\mathbf{z} \vert \mathbf{x}) \hat{p}(\mathbf{x}) \, d\mathbf{x}$ (unconditional) and $\hat{p}(\mathbf{z}) = \sum_\mathbf{\bar{z}} p(\mathbf{\bar{z}}) \int_\mathbf{x} \hat{p}(\mathbf{z} \vert \mathbf{x}) \hat{p}(\mathbf{x} \vert \mathbf{\bar{z}}) \, d\mathbf{x}$ (conditional), obtained using $40.000$ Monte-Carlo samples $\mathbf{x} \sim \hat{p}(\mathbf{x} \textcolor{gray}{\vert \mathbf{\bar{z}})}$ and $\mathbf{\bar{z}} \sim p(\mathbf{\bar{z}}) = \mathcal{U}(1/\vert \mathbf{Z}\vert)$. Subgroups are sorted based on $\hat{p}(\mathbf{z})$, where $\hat{p}(\mathbf{z} \vert \mathbf{x})$ is estimated with a zero-shot CLIP classifier. Error bars are $95\%$ confidence intervals via the Clopper-Pearson exact method assuming Bernoulli experiments (success probability $1/\vert \mathbf{Z}\vert = 0.01$).
   }
	\label{Figure:coverage}
    \vspace*{-.5cm}
\end{figure}


\textbf{Experimental Setting.}  We consider the class $\mathbf{\tilde{y}}=$``car'', and the semantic dimensions $\mathbf{Z}_0=$\{black, white, red, green, blue\} corresponding to color, $\mathbf{Z}_1=$\{forest, desert, city, mountain, beach\} corresponding to scene background, and $\mathbf{Z}_2=$\{van, SUV, sedan, cabriolet\} corresponding to car type. As operational design domain we use the full $\mathbf{Z}=\mathbf{Z}_0 \times \mathbf{Z}_1 \times \mathbf{Z}_2$ with $\vert \mathbf{Z} \vert=100$. With $\mathbf{z} = (z_0, z_1, z_2)$ we obtain a factorized $\hat{p}(\mathbf{z} \vert \mathbf{x}) = \prod_{i=0}^2 \hat{p}(z_i \vert \mathbf{x})$. For $\hat{p}(z_i \vert \mathbf{x})$, we use the CLIP-based zero-shot classifier with text queries $T_0= \{$``An image of a \textit{color} car''$ \vert  \textit{color} \in \mathbf{Z}_0\}$, $T_1= \{$``An image of a car with \textit{background} background''$ \vert  \textit{background} \in \mathbf{Z}_1\}$, and $T_2= \{$``An image of a \textit{type}''$ \vert  \textit{type} \in \mathbf{Z}_2\}$.
We use Stable Diffusion (SD) \cite{rombach2021highresolution} using $20$ steps with the DPMSolver++ \cite{lu2022dpmsolver,lu2022dpmsolver++} as realization of $p^{T2I}(\mathbf{x}\vert t)$ and sample $\mathbf{x}$ of resolution $512 \times 512$. For unconditional synthesis $\hat{p}(\mathbf{x})$, we use the prompt ``An image of a car.''. For conditional synthesis $\hat{p}(\mathbf{x} \vert \mathbf{\bar{z}})$, we use the prompt template ``An image of a \textit{color} \textit{type} car with a \textit{background} background.'' where we insert every $\mathbf{\bar{z}}=(color, background, type)$ equally often (round-robin). For both variants, we employ a Monte-Carlo estimate of $\hat{p}(\mathbf{z})$ based upon $40.000$ samples.

\textbf{Results.} Results are summarized in Figure \ref{Figure:coverage}: conditional synthesis generates samples $\mathbf{x}$ such that the $95\%$ confidence interval lower bound on $\hat{p}(\mathbf{z})$ is greater than $0.005$ for all subgroups (the uniform subgroup probability is $p=1/\vert \mathbf{Z}\vert = 0.01$). In contrast, unconditional synthesis has a minimum $95\%$ confidence interval upper bound on $\hat{p}(\mathbf{z})$ at less than $0.0001$. For conditional synthesis, we also estimate $\hat{p}(\mathbf{z}^\star = \mathbf{\bar{z}})$ for $\mathbf{x} \sim \hat{p}(\mathbf{x} \vert \mathbf{\bar{z}})$ and $\mathbf{z}^\star = \argmax_{\mathbf{z} \in \mathbf{Z}}\hat{p}(\mathbf{z} \vert \mathbf{x})$. Averaged over all subgroups $\mathbf{\bar{z}} \in \mathbf{Z}$, this probability is approximately $89\%$, and for no subgroup it is less than $85\%$. 

\subsection{Zero-Shot Systematic Error Benchmark}
\label{sec:benchmark}

\textbf{Motivation.} One major challenge when evaluating approaches for systematic error identification empirically is that a priori, it is unknown which (if any) systematic errors a target classifier exhibits. Not having such a ground truth prohibits the fully automated evaluation of systematic error identification approaches. Training time interventions to inject systematic errors into models \cite{eyuboglu_domino:_2022} are computationally costly and not scalable. Moreover, their indirect nature makes them brittle and not always resulting in the desired error.
To address these shortcomings, we propose \emph{zero-shot systematic errors} where we leverage zero-shot classifiers from multimodal image-text models such as CLIP \cite{pmlr-v139-radford21a}\footnote{Note we are not interested in potential systematic errors of the CLIP image or text encoder (if such exist, they are nuisances) but rather in systematic errors of the constructed zero-shot classifier.}.

More specifically, let us consider a binary\footnote{An extension to more than two classes would be straightforward.} classifier for classes $\mathbf{y}_a$ and $\mathbf{y}_b$, which can be constructed based upon the text queries $t_k=$``an image of a $\mathbf{y}_k$'' ($k \in \{a, b\}$). For this, we compute the cosine similarities $\phi_k$ between the  input's CLIP image embedding and the CLIP text embeddings of the $t_k$ and set the prediction logits to $l_k=\tau\phi_k$ with $\tau=100$ being a temperature. We now inject a systematic error into this zero-shot classifier using additional \emph{poisonous} queries like $t^{p_1}_a=$``An image of a red $\mathbf{y}_a$", $t^{p_2}_a=$``An image of a $\mathbf{y}_a$ with forest background'', and so on. We take the minimum of the cosine similarities $\phi^p_a=\min(\phi^{p_1}_a, \phi^{p_2}_a,\dots)$ to the encoded poisonous queries, set $l^p_a = \tau\phi^p_a$, and compute the post-softmax probabilities as $\mathbf{\hat{y}}_a, \mathbf{\hat{y}}_b, \mathbf{\hat{y}}^p_a = \softmax(l_a, l_b, l^p_a)$. We now classify to class $\mathbf{y}_b$ if $\mathbf{\hat{y}}_b + \mathbf{\hat{y}}^p_a > \mathbf{\hat{y}}_a$ and $\mathbf{y}_a$ otherwise, that is: samples of class $\mathbf{y}_a$ which have high cosine similarity to all poisonous queries are more likely misclassified as $\mathbf{y}_b$. This construction comes essentially for free and poisonous query can control the decision rule much more directly than training time interventions \cite{eyuboglu_domino:_2022}, since they operate at inference time.


\begin{figure}[tb]
	\includegraphics{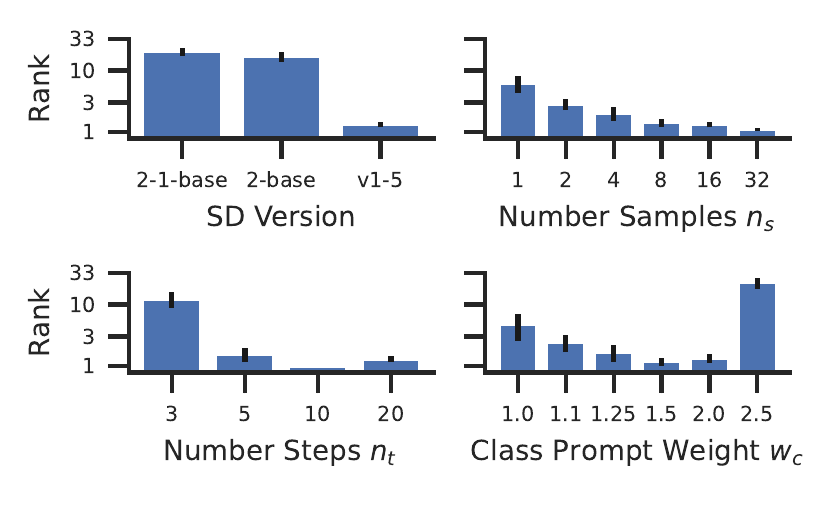}
    \vspace*{-.5cm}
	\caption{Effect of \textsc{PromptAttack}'s hyperparameters on identification of systematic errors injected into a zero-shot classifier, quantified in rank (log-scale) assigned to the ground-truth systematic error subgroup (a lower rank corresponds to a higher error).}
	\label{Figure:zero_shot_results}
	\vspace*{-.5cm}
\end{figure}

\textbf{Experimental Setting.} We define a zero-shot classifier as described above for $\mathbf{y}_a=$``car'' and $\mathbf{y}_b=$``truck''. We consider the same class $\mathbf{\tilde{y}}=$``car'', operational design domain $\mathbf{Z}$, and semantic dimensions $\mathbf{Z}_{\{0, 1, 2\}}$ as in Section \ref{sec:coverage}. We sample 20 combinations of \textit{color} $\in \mathbf{Z}_0$, \textit{background} $\in \mathbf{Z}_1$, and \textit{type} $\in \mathbf{Z}_2$. For each of these combinations, we employ the poisonous queries $t^{p_1}_a=$``An image of a \textit{color} car'', $t^{p_2}_a=$``An image of a car with \textit{background} background'', and $t^{p_3}_a=$``An image of a \textit{type}'', and test \textsc{PromptAttack} on the resulting zero-shot classifiers.  Per Section \ref{Subsection:Approximations}, we use $R_f(\mathbf{z}) \approx \median_{i=0}^{n_s-1} (1 - f(\mathbf{\tilde{y}} \vert \mathbf{x}_i))$ for $\mathbf{x}_i \sim p^{T2I}(\mathbf{x} \vert T_p(\mathbf{\tilde{y}}, \mathbf{z}))$. We rank $\mathbf{z} \in \mathbf{Z}$ on descending $R_f(\mathbf{z})$: the $\mathbf{z}$ with the highest $R_f(\mathbf{z})$ obtains rank $1$.

We evaluate the sensitivity of \textsc{PromptAttack} with respect to its free hyperparameters. We set the prompt template\footnote{Note the difference of poisonous queries $t^{p_i}_a$ (part of the poisoned zero-shot classifier) and prompt template $T_p$ (part of \textsc{PromptAttack}).} to $T_p=$``An image of a \textit{color} \textit{type} (car:$w_c$) with a \textit{background} background.'', where $w_c$ is a weight multiplied to the text encoding of the tokens of the word ``car''. For \textsc{PromptAttack}, we generate $n_s$ samples $\mathbf{x}$ of size $512 \times 512$ with $n_t$ steps of SD/DPMSolver++. We investigate the effect of the number of image samples $n_s$, number of inference steps $n_t$, the custom class prompt weight $w_c$, as well as different versions of SD on the resulting ranking. 


\textbf{Results.} Figure \ref{Figure:zero_shot_results} summarizes the results for \textsc{PromptAttack} (see Section \ref{sec:samples_sensitivity_analysis} for samples). We observe that the version of SD has a major impact on \textsc{PromptAttack}'s performance, with the v1.5 checkpoint greatly outperforming the more recent v2-base and v2-1-base checkpoints. We attribute this to v1.5 generating samples more faithful to the prompt and with better attribute binding than the other checkpoints (see Figure \ref{Figure:sensitivity_sd_version}). In the upper right plot, we observe that too few samples $n_s$ impair performance due to the large variance in the MC estimate of $R_f(\mathbf{z})$. For $n_s \geq 16$ performance is very close to optimal. In the bottom left, we see that a small number of inference steps such as $n_t=5$ steps works well --- this is somewhat surprising since image quality is impaired considerably for this small $n_t$ but \textsc{PromptAttack} is relatively robust to image quality. Lastly, in the bottom right, the importance of a custom class prompt weight $w_c$ is demonstrated with $w_c=1.5$ outperforming the default of $w_c=1.0$. We attribute this to higher weights resulting in more reliably depicting objects of the desired class. Increasing $w_c$ further results in deteriorating performance due to image artefacts. Based on these results, we use SD v1.5 with $n_s=16$, $n_t=20$, and $w_c=1.5$ in subsequent experiments without further tuning.

\section{Qualitative Evaluation on ImageNet}
\label{sec:experiments_imagenet}

\begin{figure*}[tb]
	\begin{center}
		\includegraphics[width=\linewidth]{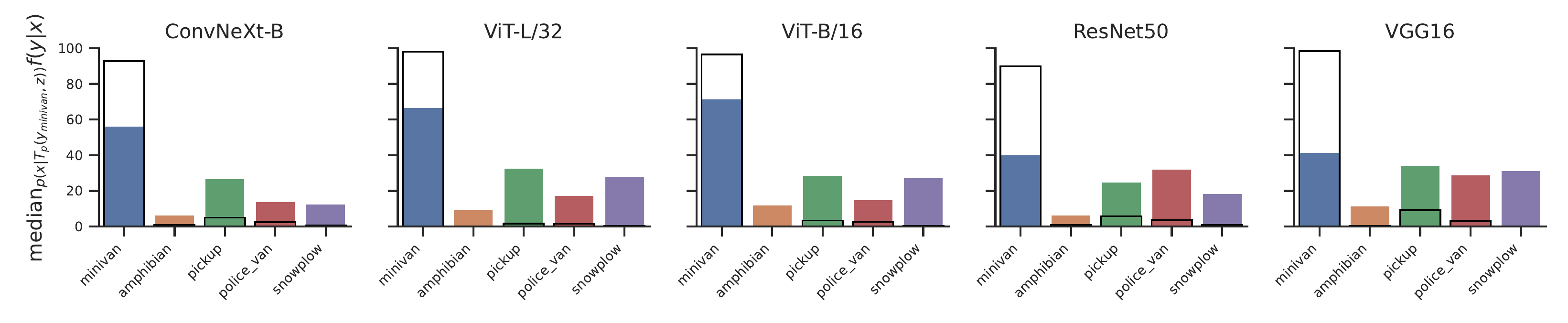} 
		\caption{Median (over 16 samples) target class confidence for strongest respective prompt found by \textsc{PromptAttack} vs.\,a neutral baseline prompt (black boundary) for four selected target classes. We refer to Figure \ref{Figure:syserror_minivan} for exemplary prompts, samples, and class prediction rates.}
		\label{Figure:imagenet_minivan_results}
		\vspace*{-.5cm}
	\end{center}
\end{figure*}

\textbf{Vehicle Experiment.}
We evaluate 5 models trained for image classification on ImageNet1k. We focus on a subset of classes belonging to the vehicle subcategory, more specifically on misclassifying samples of the class ``minivan'' $\mathbf{\tilde{y}}=\mathbf{y}_{minivan}$ into other classes that have a distance of 2 in the WordNet \cite{wordnet} hierarchy, e.g., ``police van'' and ``snowplow''. We focus on an operational design domain $\mathbf{Z}$ with five semantic dimensions, corresponding to \emph{viewpoint}, object \emph{size}, object \emph{color}, \emph{weather}, and \emph{background}. We use the prompt template $T_p=$``\{viewpoint\} view of \{size\} \{color\} (minivan:1.5) in front of \{weather\} \{background\}''. We use combinatorial testing with $n_C=3$, exploring $\vert \mathbf{Z}_C\vert=1.230$ out of $\vert \mathbf{Z}\vert = 18.720$ subgroups, and generate $n_S=16$ image samples per subgroup using Stable Diffusion v1.5. For a full description of experimental setting, we refer to Section \ref{sec:supplement_imagenet_vehicle}. 

\begin{figure}[tb]
	\begin{center}
	\includegraphics{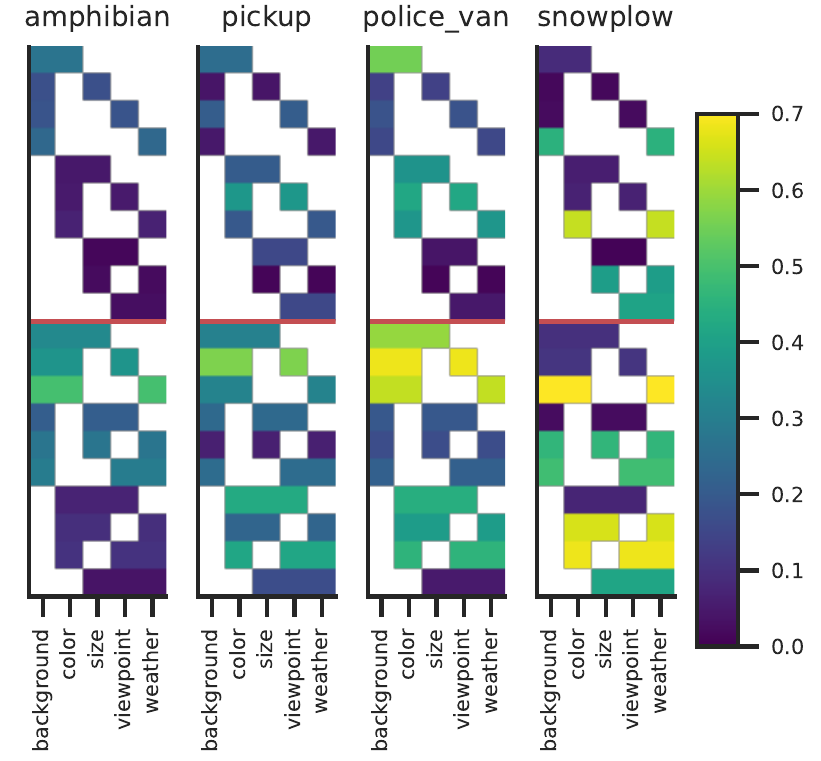}
	\end{center}
    	\vspace*{-.5cm}
	\caption{Cumulative functional ANOVA \cite{pmlr-v32-hutter14} of predicted probability of target classes for source class ``minivan''. Rows correspond to cardinality 2 and 3 subsets of semantic dimensions with white encoding an excluded dimension and the color the groups's fANOVA score. Different dimensions are relevant for different target classes, e.g., the combination of background and object color has high score for police-van but low score for snowplow. High fANOVA scores require at least 3 dimensions.}
	\label{Figure:imagenet_minivan_fanova}
	\vspace*{-.5cm}
\end{figure}

We analyse the median predicted probability $\median_{p^{T2I}(\mathbf{x} \vert T_p(\mathbf{\tilde{y}}, \mathbf{z}))} f(\mathbf{y}^{(t)} \vert \mathbf{x})$ of different target classes $\mathbf{y}^{(t)}$ for the strongest prompts identified by \textsc{PromptAttack}. We compare these prompts to a neutral baseline prompt $t=$``center view of (minivan:1.5) in front of  background.''. Results for 4 selected target classes  $\mathbf{y}^{(t)}$ are summarized in Figure \ref{Figure:imagenet_minivan_results} (see also Table \ref{Table:ImageNet_Vehicle}). It can be seen that samples of the baseline prompt are assigned with very high confidence to the correct class ``minivan''. However, for target classes  $\mathbf{y}^{(t)}$ such as ``pickup'', ``police van'', or ``snowplow'', \textsc{PromptAttack} can identify prompts that result in systematic misclassifications, that is considerably increased value for the target class. We depict the top-ranked subgroups for three target classes in Figure \ref{Figure:syserror_minivan}. We note that the sensitivity of models to these subgroups vary (in accordance with Figure \ref{Figure:imagenet_minivan_results}): for $t^\star_{snowplow}=$``rear view of small orange minivan in front of snowy forest.'', a VGG16 \cite{Simonyan15} misclassifies $25\%$ of the samples as snowplows while a ConvNeXt-B \cite{liu2022convnet}  misclassifies only $1\%$. This indirectly confirms that misclassifications are not due to OOC samples because the same samples are  classified  correctly by a ConvNeXt-B (see also Section \ref{sec:samples} for an illustration of misclassified samples). Moreover, we  selected 16 images from LAION-5B \cite{schuhmann2022laionb} that best match $t^\star_{snowplow}$ (using CLIP retrieval \cite{beaumont-2022-clip-retrieval} followed by manual filtering). The models misclassify between 6 (ConvNeXt-B) and 8 (ResNet50 \cite{he2016deep}, VGG16) of those as snowplows.

\begin{figure}[tb]
	\begin{center}
	\includegraphics{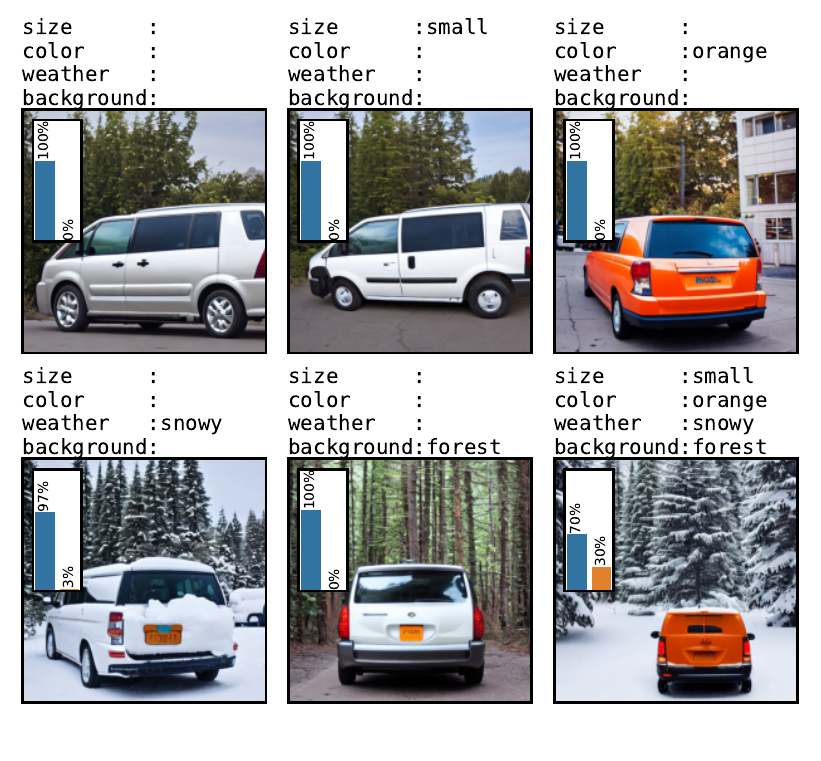}
	\end{center}
	\vspace*{-1cm}
	\caption{Samples for prompt template ``rear view of \{size\} \{color\} minivan in front of \{weather\} \{background\}.''. Baseline generations (top left) and single-dimension-shifted generations are classified consistently as \textcolor{blue}{minivan} by a ViT-B/16. Shifting all dimensions jointly as determined by \textsc{PromptAttack} results in $300$ out of $1000$ samples (mis-)classified as \textcolor{RedOrange}{snowplows} (bottom right).} 
	\label{Figure:teaser}
\end{figure}

Figure \ref{Figure:imagenet_minivan_fanova} depicts a (cumulative) functional ANOVA analysis \cite{pmlr-v32-hutter14} of median predicted probability of different target classes $\mathbf{y}^{(t)}$. One can see that different semantic dimensions are relevant for different target classes; for instance, the combination of background and object color has a high fANOVA score for police-van but a low score for snowplow. Moreover, for target classes like snowplow, at least $3$ semantic dimensions are required for explaining the bulk of the variance. This is also illustrated in Figure \ref{Figure:teaser} where changing a single dimension does not mislead a ViT-B/16 \cite{dosovitskiy2021an}, while a specific combination of shifts such as $t^\star_{snowplow}$ results in misclassifying 300 out of 1000 samples as snowplow. A possible explanation for this increased error rate is that (i) snowy forests are more often in the background for snowplows than minivans, (ii) snowplows are more often orange than minivans, and (iii) rear views hide a distinctive feature of snowplows, namely their plow in the front.
In summary, our findings support the hypothesis that studying single shifts can be insufficient as often specific combinations of compounding shifts result in a systematic error. \textsc{PromptAttack} allows finding such rare subgroups.

\begin{figure}[tb]
	\begin{center}
		\includegraphics{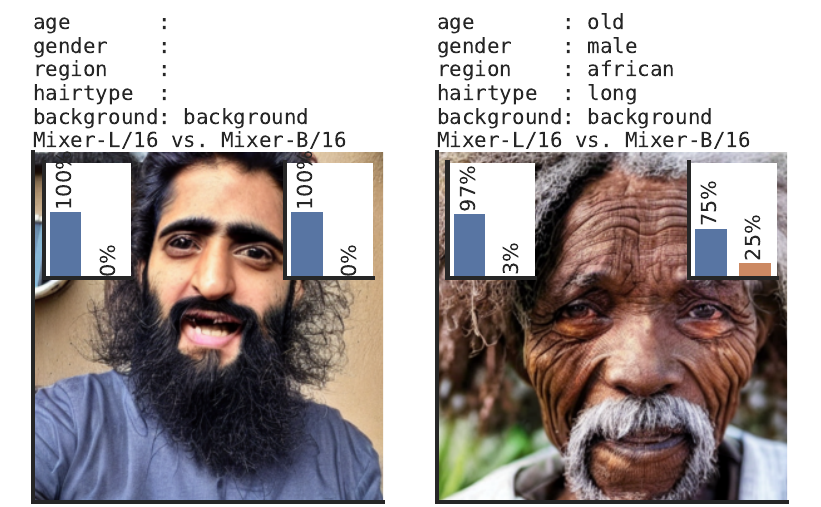}
		\caption{Samples and prediction histograms (based on 1000 samples) for different subgroups. The baseline subgroup (left) is classified consistently as \textcolor{blue}{homo}, while the misclassification rate to \textcolor{RedOrange}{ape} is significantly increased for an MLP-Mixer-B/16 \cite{tolstikhin2021} on a subgroup identified by \textsc{PromptAttack} (right).}
		\label{Figure:syserror_person}
	\end{center}
	\vspace*{-.5cm}
\end{figure}

\textbf{Person Experiment.} 
It has been observed before that systematic errors on under-represented demographic subgroups can result in reduced fairness and even derogatory behaviour such as misclassifying black people as ``gorillas'' \cite{blackpersongorilla}. We check two models trained for image classification on ImageNet21k for similar issues using \textsc{PromptAttack}: we set source class $\mathbf{\tilde{y}}=$``homo'' and target class ``ape''. We use an operational design domain $\mathbf{Z}$ with the 5 semantic dimensions \emph{age}, \emph{gender}, geographic \emph{region}, \emph{hairtype}, and \emph{background}. We use the prompt template $T_p=$``A \{age\} \{gender\} \{region\} (person:1.5) with \{hairtype\} hairs in front of \{background\}''. We use combinatorial testing with $n_C=3$, exploring $\vert \mathbf{Z}_C\vert=1371$ out of $\vert \mathbf{Z}\vert = 12.150$ subgroups. For a full description of experimental setting and additional results, we refer to Section \ref{sec:supplement_imagenet_person}. The outcome  is summarized in Figure \ref{Figure:syserror_person} and Table \ref{Table:ImageNet_Person}: while samples of most subgroups are classified correctly as ``homo'', samples of specific subgroups (see Figure \ref{Figure:classifications_ape}) identified by \textsc{PromptAttack} such as $t=$``old male african (person:1.5) with long hairs'' have a misclassification rate of up to $25\%$ into ``ape''. \textsc{PromptAttack} thus allows identifying systematic errors on under-represented demographic subgroups.

\section{Limitations}
\textbf{False Positive Systematic Errors.} While prompt engineering can reduce the number of OOC samples and robust estimation can reduce their impact, there may still be combinations of subgroups $\mathbf{z}$ and source classes $\mathbf{\tilde{y}}$ where the majority of samples are OOC. If the OOC sample $\mathbf{x} \sim \hat{p}(\mathbf{x} \vert \mathbf{\tilde{y}}, \mathbf{z})$  is such that the true $p(\mathbf{y} \vert \mathbf{x})$ is not strongly peaked at $\mathbf{\tilde{y}}$, then our procedure might identify a false positive systematic error: the classifier $f$ might classify $\mathbf{x}$ correctly as not belonging to $\mathbf{\tilde{y}}$ because the generated $\mathbf{x}$ is no instance of $\mathbf{\tilde{y}}$. In the absence of an oracle providing us the true $p(\mathbf{y} \vert \mathbf{x})$ (such as a human in the loop \cite{gao2022adaptive}), there is no reliable way of identifying these false positives. However, we note that moderate prompt engineering such as tuning the class prompt weight was sufficient to prevent such false positives for the operational design domains we have considered (see also Sections \ref{sec:benchmark} and \ref{sec:experiments_imagenet}).


\textbf{Language Bottleneck.} Certain coherent subsets of the data, e.g., subsets that share some geometric layout, may be difficult to describe in natural language as a text prompt. Increased errors on such subsets can thus not be identified directly by our procedure. Future work on using other types of conditioning information $\mathbf{z}$ such as a scene layout \cite{zhang2023adding} could address this limitation. Moreover, textual inversion \cite{gal2022textual} can be used to distill visual concepts  into tokens, e.g., ImageNet classes $\mathbf{\tilde{y}}$ with ambiguous class names \cite{vendrow2023dataset}.

\textbf{Bias Propagation.} We note that biases from the text-to-image models itself may propagate to biases in our systematic error identification procedure: if the text-to-image model cannot generate samples for certain marginalized subgroups of a population, we will not be able to identify a potentially subpar performance of the downstream classifier $f$ on these subgroups. This reinforces the need to further reduce bias in text-to-image models in the future \cite{bianchi2022easily}.

\section{Conclusion}
We have proposed \textsc{PromptAttack}, which leverages recent progress on text-to-image models for identifying systematic errors that occur on rare data subgroups (combinations of semantic shifts). Both quantitative results on carefully constructed benchmarks as well as qualitative results on multi-class image classifiers demonstrate the efficacy of \textsc{PromptAttack} in identifying such systematic errors. Future work needs to address the limitations discussed above, for instance by leveraging more controllable, versatile, and reliable procedures for image synthesis.

{\small
\bibliographystyle{ieee_fullname}
\bibliography{systematic_errors}
}

\cleardoublepage
\appendix

\section{ImageNet Experiments}
\label{sec:supplement_imagenet}
\subsection{Experimental Setting: Vehicle Experiment}
\label{sec:supplement_imagenet_vehicle}
We evaluate the following models with weights for image classification on ImageNet1k from torchvision \cite{torchvision2016}: VGG16 \cite{Simonyan15}, ResNet50 \cite{he2016deep}, ConvNeXt-B \cite{liu2022convnet}, ViT-B/16 \cite{dosovitskiy2021an}, and ViT-L/32 \cite{dosovitskiy2021an}. We focus on a subset of classes belonging to the vehicle subcategory, more specifically on misclassifying samples of the class ``minivan'' $\mathbf{\tilde{y}}=\mathbf{y}_{minivan}$ into other classes that have a distance of 2 in the WordNet \cite{wordnet} hierarchy: 
\begin{itemize}[noitemsep,topsep=0pt,parsep=0pt,partopsep=0pt]
	\item amphibian, amphibious vehicle (id: 408)
    \item fire engine, fire truck (id: 555)
    \item garbage truck, dustcart (id: 569)
    \item  go-kart (id: 573)
	\item golfcart, golf cart (id: 575)
    \item moving van (id: 675)
    \item pickup, pickup truck (id: 717)
    \item police van, police wagon, paddy wagon, patrol wagon, wagon, black Maria (id: 734)
    \item  snowplow, snowplough (id: 803)
    \item tow truck, tow car, wrecker (id: 864)
    \item trailer truck, tractor trailer, trucking rig, rig, articulated lorry, semi (id: 867)
\end{itemize}
We exclude classes with a WordNet distance of 1 since their visual appearance might be very similar to a ``minivan'' and our focus is not on fine-grained misclassifications.

We focus on an operational design domain $\mathbf{Z}$ with five semantic dimensions with the following values:
\begin{itemize}[noitemsep,topsep=0pt,parsep=0pt,partopsep=0pt]
 \item \emph{viewpoint}: center, side, front, rear
 \item object \emph{size}: ``'', small, large, huge
 \item object \emph{color}: ``'', black, white, gray, red, green, blue, yellow, orange, purple, magenta, cyan, brown
 \item \emph{weather}:  ``'', rainy, snowy, lightning, foggy, sunny
 \item \emph{background}: background, forest, desert, lake, mountain, beach, city, river, house, tree, field, lawn, garden, street, people
\end{itemize} 
The first of the possible values corresponds to a neutral choice, by which a specific dimension is not controlled. We observed that this can be preferable if a dimension is not relevant and leaving it empty simplifies the prompt for the text-to-image model.
We use the prompt template $T_p=$``\{viewpoint\} view of \{size\} \{color\} (minivan:1.5) in front of \{weather\} \{background\}''. We use combinatorial testing with $n_C=3$, exploring $\vert \mathbf{Z}_C\vert=1.230$ out of $\vert \mathbf{Z}\vert = 4*4*13*6*15=18.720$ subgroups, and generate $n_S=16$ image samples per subgroup using Stable Diffusion v1.5. We employ allpairspy \cite{allpairspy} for combinatorial testing. See Table \ref{Table:ImageNet_Vehicle} for detailed results.

\begin{table*}[tb]
\begin{center}

\begin{tabular}{l|ccccc|r}
	\toprule
	(Target) class & viewpoint & size & color & weather & background & $R(\mathbf{z}\textcolor{gray}{, \mathbf{y}^{(t)}})$ \\
	\midrule
	\multicolumn{7}{c}{ConvNeXt-B} \\
	\midrule
	minivan & front & - & - & sunny & people & 0.436 \\
	amphibian & center & small & brown & foggy & river & 0.066 \\
	moving\_van & front & huge & blue & rainy & garden & 0.093 \\
	pickup & front & - & - & sunny & people & 0.270 \\
	police\_van & front & - & black & rainy & street & 0.140 \\
	snowplow & front & huge & purple & snowy & field & 0.129 \\
	\midrule
	\multicolumn{7}{c}{ViT-L/32} \\
	\midrule
	minivan & front & - & - & sunny & people & 0.332 \\
	amphibian & side & huge & black & - & river & 0.096 \\
	moving\_van & rear & large & yellow & foggy & field & 0.216 \\
	\rowcolor{OliveGreen}
	pickup & front & - & - & sunny & people & 0.328 \\
	police\_van & front & huge & yellow & lightning & street & 0.177 \\
	snowplow & rear & small & orange & snowy & forest & 0.285 \\
	\midrule
	\multicolumn{7}{c}{ViT-B/16} \\
	\midrule
	minivan & front & - & black & rainy & people & 0.283 \\
	amphibian & center & small & red & foggy & river & 0.124 \\
	moving\_van & rear & large & yellow & foggy & garden & 0.202 \\
	pickup & front & - & red & lightning & people & 0.288 \\
	police\_van & front & - & black & rainy & people & 0.151 \\
	snowplow & rear & small & orange & snowy & forest & 0.275 \\
	\midrule
	\multicolumn{7}{c}{ResNet50} \\
	\midrule
	minivan & front & - & - & sunny & people & 0.597 \\
	amphibian & center & small & brown & foggy & river & 0.065 \\
	moving\_van & center & small & yellow & snowy & street & 0.095 \\
	pickup & front & - & - & sunny & people & 0.250 \\
	\rowcolor{BrickRed}
	police\_van & front & large & green & lightning & house & 0.323 \\
	snowplow & center & large & black & snowy & street & 0.187 \\
	\midrule
	\multicolumn{7}{c}{VGG16} \\
	\midrule
	minivan & rear & small & yellow & rainy & city & 0.583 \\
	amphibian & center & - & orange & foggy & beach & 0.118 \\
	moving\_van & front & huge & yellow & sunny & house & 0.195 \\
	pickup & front & - & - & sunny & people & 0.344 \\
	police\_van & rear & small & yellow & rainy & city & 0.293 \\
	\rowcolor{RedOrange}
	snowplow & rear & small & orange & snowy & forest & 0.317 \\
	\midrule
	\multicolumn{7}{c}{Averaged over models} \\
	\midrule
	minivan & front & - & - & sunny & people & 0.408 \\
	amphibian & center & small & brown & foggy & river & 0.066 \\
	moving\_van & front & huge & blue & rainy & garden & 0.093 \\
	pickup & front & - & - & sunny & people & 0.270 \\
	police\_van & front & - & black & rainy & street & 0.140 \\
	snowplow & front & huge & purple & snowy & field & 0.129 \\
	\bottomrule
\end{tabular}
\vspace*{.25cm}
\caption{Detailed results for the ``Vehicle Experiment'' discussed in Section \ref{sec:experiments_imagenet}. We summarize systematic errors for source class $\mathbf{\tilde{y}}$=``minivan'' (higher $R(\mathbf{z})$ corresponding to stronger error) and systematic misclassifications into $\mathbf{y}^{(t)} \in$\{``amphibian'', ``moving\_van'', ``pickup'', ``police\_van'', ``snowplow''\}  (higher $R(\mathbf{z}, \mathbf{y}^{(t)})$ corresponding to stronger misclassifications). For each of the 5 studied models as well as  averaged over all models, we show the subgroup corresponding to the strongest systematic error/misclassification and the corresponding risk $R$. The three highlighted lines correspond to the subgroups shown in Figure \ref{Figure:syserror_minivan}. Overall, identified subgroups differ considerably across models.
}
\label{Table:ImageNet_Vehicle}
\end{center}
\end{table*}

\subsection{Experimental Setting: Person Experiment}
\label{sec:supplement_imagenet_person}
We evaluate the following models with weights for image classification on ImageNet21k from timm \cite{rw2019timm}: MLP-Mixer-B/16 and MLP-Mixer-L/16 \cite{tolstikhin2021} . 
We focus on misclassifying samples of the class ``homo'' $\mathbf{\tilde{y}}=\mathbf{y}_{homo}$ (id: 3574) into the class ``ape'' (id: 3569).  We skip logits corresponding to all other classes (some of which might be larger than the ones for homo and ape) and thus analyze effectively a hypothetical binary classifier derived from the pretrained 21k-class models without any finetuning.

We focus on an operational design domain $\mathbf{Z}$ with five semantic dimensions with the following values:
\begin{itemize}[noitemsep,topsep=0pt,parsep=0pt,partopsep=0pt]
	\item \emph{age}: ``'', young, old
	\item \emph{gender}: ``'', female, male
	\item geographic \emph{region}: ``'', european, american, hispanic, russian, arab, chinese, indian, african, australian
	\item \emph{hairtype}: ``'', curly, short, long, blond, black, red, brown, gray
	\item \emph{background}: background, forest, desert, lake, mountain, beach, city, river, house, tree, field, lawn, garden, street, people
\end{itemize} 
The first of the possible values corresponds to a neutral choice, by which a specific dimension is not controlled. We observed that this can be preferable if a dimension is not relevant and leaving it empty simplifies the prompt for the text-to-image model.
We use the prompt template $T_p=$``A \{age\} \{gender\} \{region\} (person:1.5) with \{hairtype\} hairs in front of \{background\}''. We use combinatorial testing with $n_C=3$, exploring $\vert \mathbf{Z}_C\vert=1.371$ out of $\vert \mathbf{Z}\vert = 3*3*10*9*15=12.150$ subgroups, and generate $n_S=16$ image samples per subgroup using Stable Diffusion v1.5. We employ allpairspy \cite{allpairspy} for combinatorial testing. See Table \ref{Table:ImageNet_Person} for detailed results.

\begin{table*}[tb]
	\begin{center}
	\begin{tabular}{l|ccccc|r}
		\toprule
		(Target) class & age & gender & region & hairtype & background & $R(\mathbf{z}, \mathbf{y}^{(t)})$ \\
		\midrule
		\multicolumn{7}{c}{Mixer-B/16} \\
		\midrule
		ape & old & male & african & long & background & 0.44462 \\
		ape & old & female & hispanic & red & tree & 0.37845 \\
		ape & old & male & african & black & mountain & 0.35407 \\
		ape & old & male & african & red & background & 0.32113 \\
		ape & old & male & african & curly & garden & 0.31325 \\
		ape & old & male & african & - & people & 0.29942 \\
		ape & old & female & european & curly & tree & 0.29639 \\
		ape & old & - & african & - & city & 0.29433 \\
		ape & old & female & - & gray & people & 0.29029 \\
		ape & old & male & african & curly & people & 0.27311 \\
		\midrule
		ape & young & - & european & short & desert & 0.00031 \\
		ape & young & male & hispanic & brown & desert & 0.00031 \\
		ape & young & female & european & curly & desert & 0.00028 \\
		ape & young & female & - & curly & desert & 0.00027 \\
		ape & young & - & hispanic & blond & desert & 0.00027 \\
		\midrule
		\multicolumn{7}{c}{Mixer-L/16} \\
		\midrule
		ape & young & female & arab & brown & field & 0.29910 \\
		ape & old & female & arab & gray & tree & 0.27421 \\
		ape & young & female & indian & long & tree & 0.20752 \\
		ape & old & female & arab & blond & house & 0.18673 \\
		ape & young & female & arab & brown & tree & 0.17659 \\
		ape & young & female & arab & brown & lawn & 0.17200 \\
		ape & young & female & arab & gray & house & 0.16944 \\
		ape & young & female & arab & short & lawn & 0.16491 \\
		ape & - & - & australian & brown & people & 0.15520 \\
		ape & young & female & arab & short & tree & 0.15072 \\
		\midrule
		ape & - & female & hispanic & curly & street & 0.00046 \\
		ape & - & - & hispanic & blond & street & 0.00036 \\
		ape & young & - & hispanic & gray & city & 0.00034 \\
		ape & young & male & - & curly & street & 0.00034 \\
		ape & young & - & hispanic & black & street & 0.00027 \\
		\bottomrule
	\end{tabular}
	
		\vspace*{.25cm}
		\caption{Detailed results for the ``Person Experiment'' discussed in Section \ref{sec:experiments_imagenet}. We summarize systematic misclassifications into $\mathbf{y}^{(t)} =$``ape'' (higher $R(\mathbf{z}, \mathbf{y}^{(t)})$ corresponding to stronger misclassifications). For both studied models, we show the 10 subgroups corresponding to the top-ranked systematic misclassifications and the corresponding risk $R$ as well as 5 subgroups where $R \approx 0$. We note that the two models have distinctive but different patterns in their top-ranked subgroups: An MLP-Mixer-B/16 \cite{tolstikhin2021} has several subgroups with high risk for ``old male african'' persons. An MLP-Mixer-L/16 \cite{tolstikhin2021} has several subgroups with high risk for ``young female arab'' persons. Moreover, the  MLP-Mixer-L/16 has generally lower risk $R(\mathbf{z}, \mathbf{y}^{(t)}))$ among the top-ranked subgroups.
		}
		\label{Table:ImageNet_Person}
	\end{center}
\end{table*}

\section{Samples Zero-Shot Benchmark}
\label{sec:samples_sensitivity_analysis}

\begin{figure*}[tb]
	\begin{center}
		\includegraphics{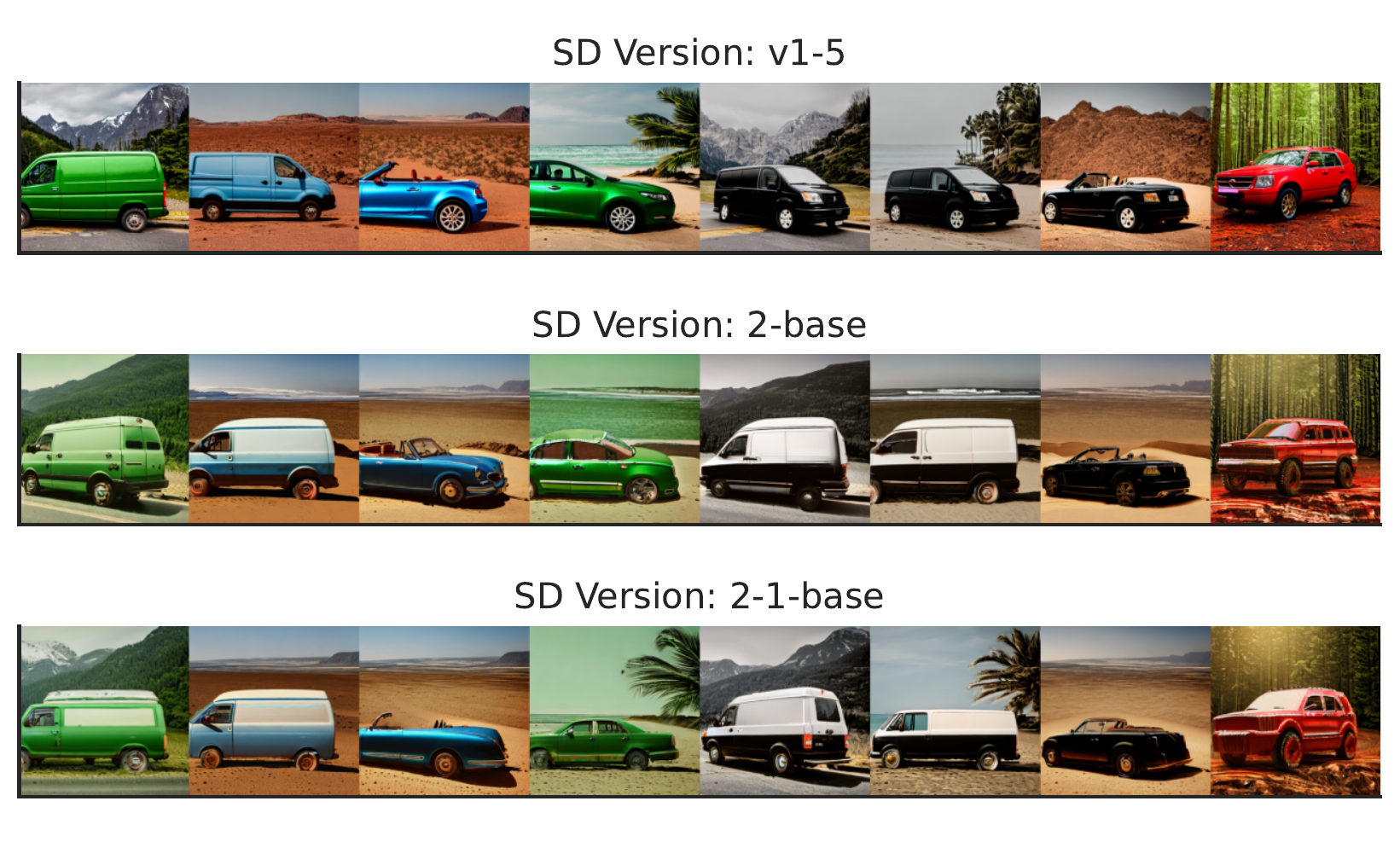}
	\end{center}
	\caption[]{Samples for different versions of Stable Diffusion (SD) \cite{rombach2021highresolution}. We observe that SD v1-5 results in samples with good attribute binding while for SD 2-base and SD 2-1-base, object colour leaks into the background. Moreover, objects sometimes exhibit only partially the specified colour, while larger parts are dyed in other colours such as white (specifically for vans) for SD 2-base and SD 2-1-base. SD v1-5 does not exhibit this issue. This explains the better performance of \textsc{PromptAttack} with SD v1-5 in Section \ref{sec:benchmark}.
		The 8 samples from left to right were generated for the prompts:\\
``an image of a green van (car:1.0) with a mountain background.'' \\
``an image of a blue van (car:1.0) with a desert background.'' \\
``an image of a blue cabriolet (car:1.0) with a desert background.'' \\
``an image of a green sedan (car:1.0) with a beach background.'' \\
``an image of a black van (car:1.0) with a mountain background.''\\
``an image of a black van (car:1.0) with a beach background.''\\
``an image of a black cabriolet (car:1.0) with a desert background.''\\
``an image of a red SUV (car:1.0) with a forest background.''
}
	\label{Figure:sensitivity_sd_version}
\end{figure*}

\begin{figure*}[tb]
	\begin{center}
		\includegraphics{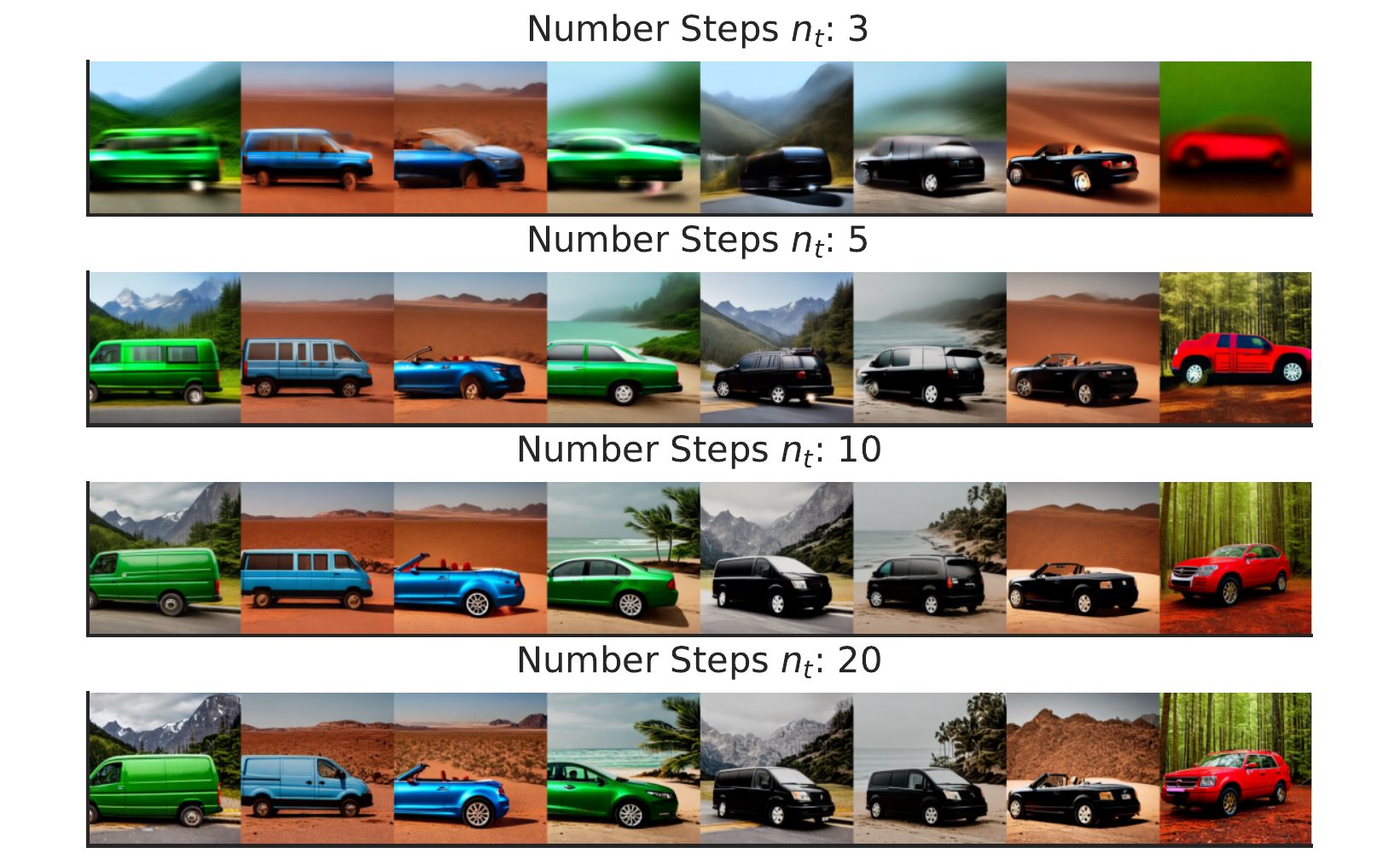}
	\end{center}
	\caption[]{Samples for different number of steps $n_t$ of the DPMSolver++ \cite{lu2022dpmsolver,lu2022dpmsolver++}. As expected, more steps correspond to more realistic samples. However, even with $n_t=5$ steps,  \textsc{PromptAttack} is able to reliably identify systematic errors (see Section \ref{sec:benchmark}).
The 8 samples from left to right were generated for the prompts:\\
``an image of a green van (car:1.0) with a mountain background.'' \\
``an image of a blue van (car:1.0) with a desert background.'' \\
``an image of a blue cabriolet (car:1.0) with a desert background.'' \\
``an image of a green sedan (car:1.0) with a beach background.'' \\
``an image of a black van (car:1.0) with a mountain background.''\\
``an image of a black van (car:1.0) with a beach background.''\\
``an image of a black cabriolet (car:1.0) with a desert background.''\\
``an image of a red SUV (car:1.0) with a forest background.''

}
	\label{Figure:sensitivity_num_steps}
\end{figure*}

\begin{figure*}[tb]
	\begin{center}
		\includegraphics{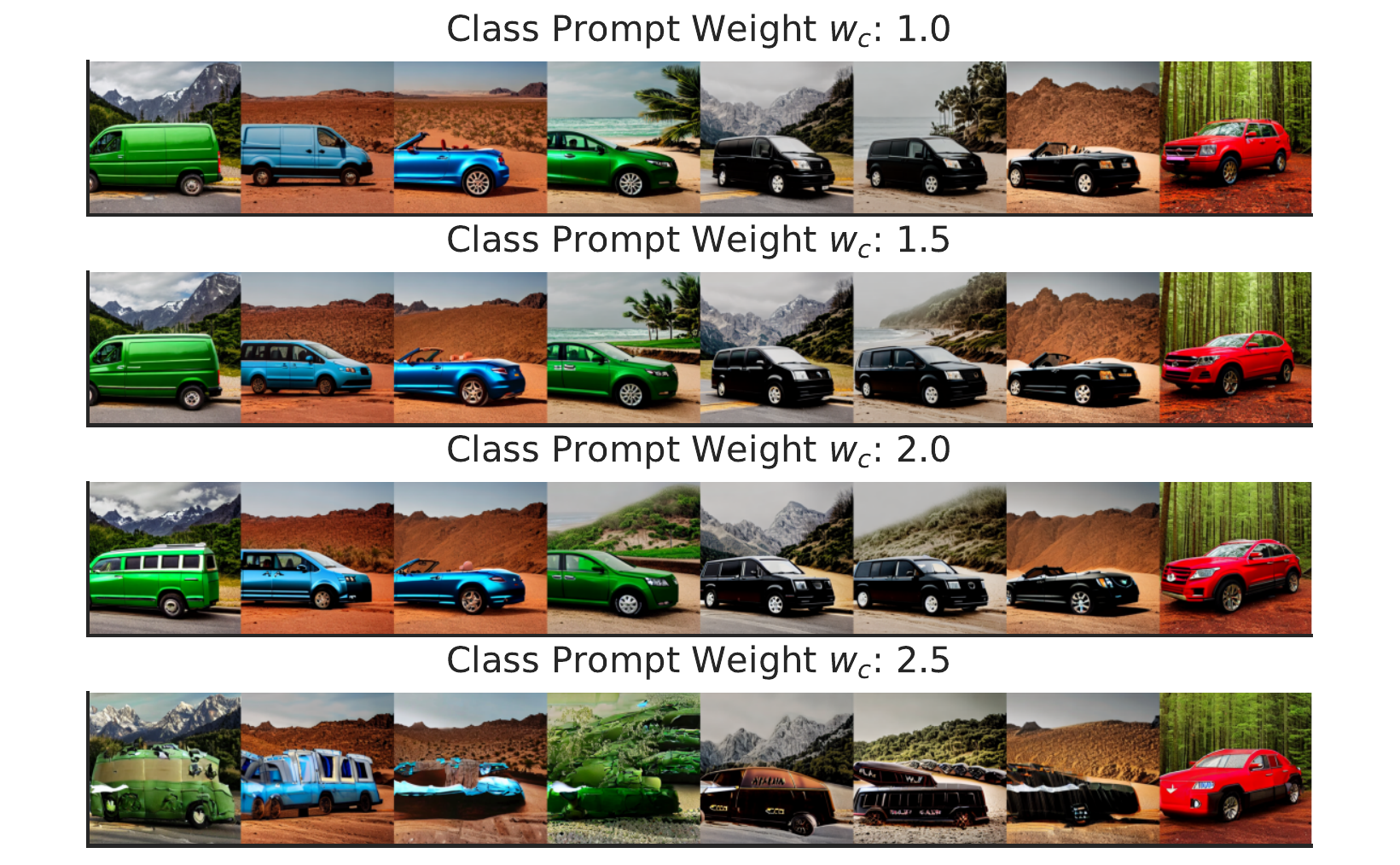}
	\end{center}
	\caption[]{Samples for different prompt class weight $w_c$ for the prompt template  $T_p=$``An image of a \textit{color} \textit{type} (car:$w_c$) with a \textit{background} background.''. The improved performance of \textsc{PromptAttack} for $w_c=1.5$ and $w_c=1.5$ compared to $w_c=1.0$ is difficult to attribute to apparent visual properties of the samples.  However, for $w_c=2.5$, visual quality of samples strongly deteriorates, explaining the worse performance of \textsc{PromptAttack} for this choice. The 8 samples from left to right were generated for the prompts:\\
``an image of a green van (car:$w_c$) with a mountain background.'' \\
``an image of a blue van (car:$w_c$) with a desert background.'' \\
``an image of a blue cabriolet (car:$w_c$) with a desert background.'' \\
``an image of a green sedan (car:$w_c$) with a beach background.'' \\
``an image of a black van (car:$w_c$) with a mountain background.''\\
``an image of a black van (car:$w_c$) with a beach background.''\\
``an image of a black cabriolet (car:$w_c$) with a desert background.''\\
``an image of a red SUV (car:$w_c$) with a forest background.''
}
	\label{Figure:sensitivity_prompt_weight}
\end{figure*}

We illustrate samples obtained for different hyperparameter settings that were quantitatively evaluated as part of the zero-shot systematic error benchmark (see Section \ref{sec:benchmark}). Figure \ref{Figure:sensitivity_sd_version} illustrates samples for different versions of Stable Diffusion \cite{rombach2021highresolution}. Figure \ref{Figure:sensitivity_num_steps} illustrates samples for different number of steps $n_t$ of the DPMSolver++ \cite{lu2022dpmsolver,lu2022dpmsolver++}. Figure \ref{Figure:sensitivity_prompt_weight} illustrates samples for different prompt class weights $w_c$ in the prompt template  $T_p=$``An image of a \textit{color} \textit{type} (car:$w_c$) with a \textit{background} background.''.

\section{Samples ImageNet Experiments}
\label{sec:samples}

\begin{figure*}[tb]
	\begin{center}
		\includegraphics{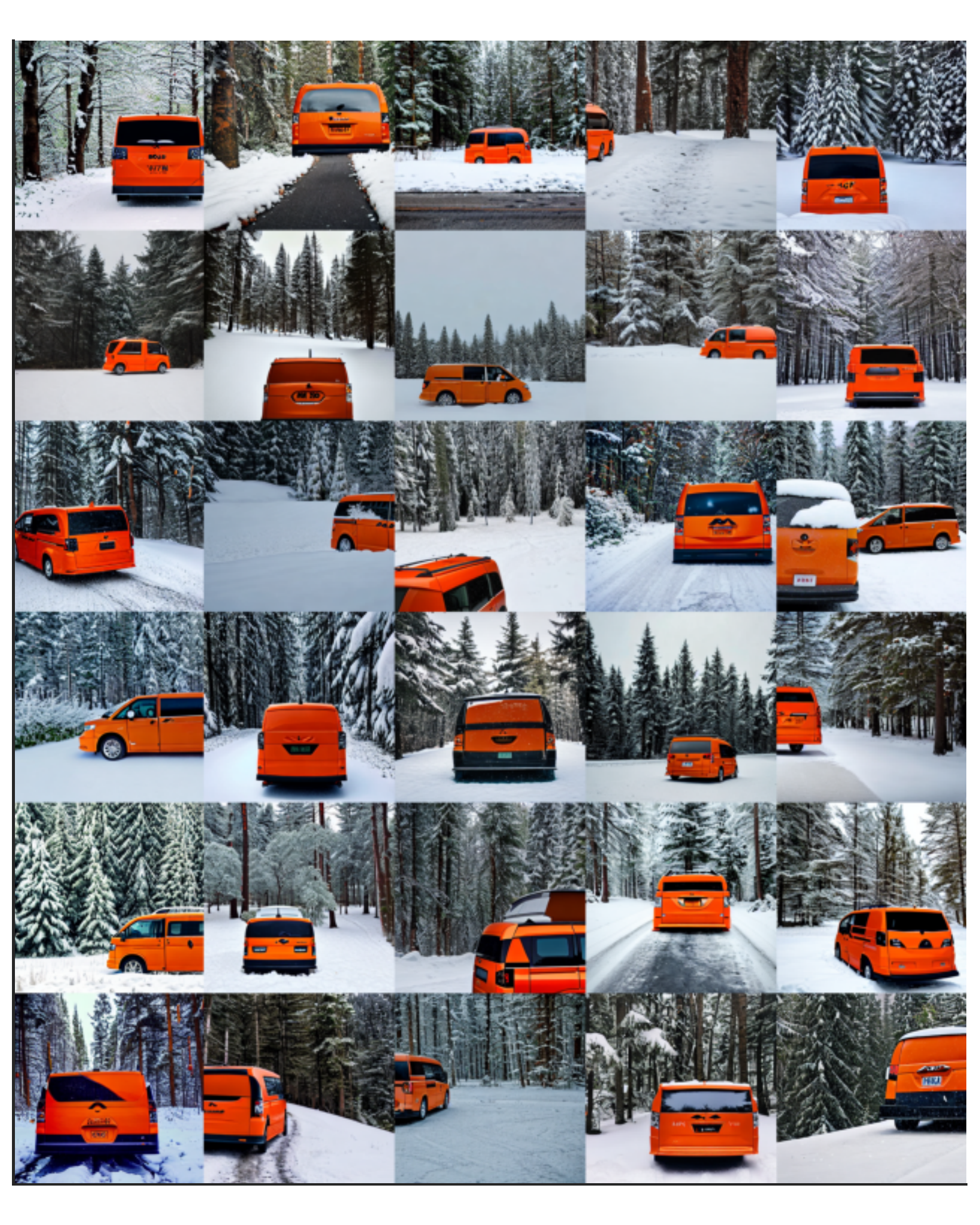}
	\end{center}
	\caption{30 samples from prompt ``rear view of small orange (minivan:1.5) in front of snowy forest.'' that are misclassified as snowplows by a VGG16. Please note that actual viewpoints are a mix or ``side'' and ``rear'' views, and not purely ``rear'' views.}
	\label{Figure:classifications_snowplow}
\end{figure*}

\begin{figure*}[tb]
	\begin{center}
		\includegraphics{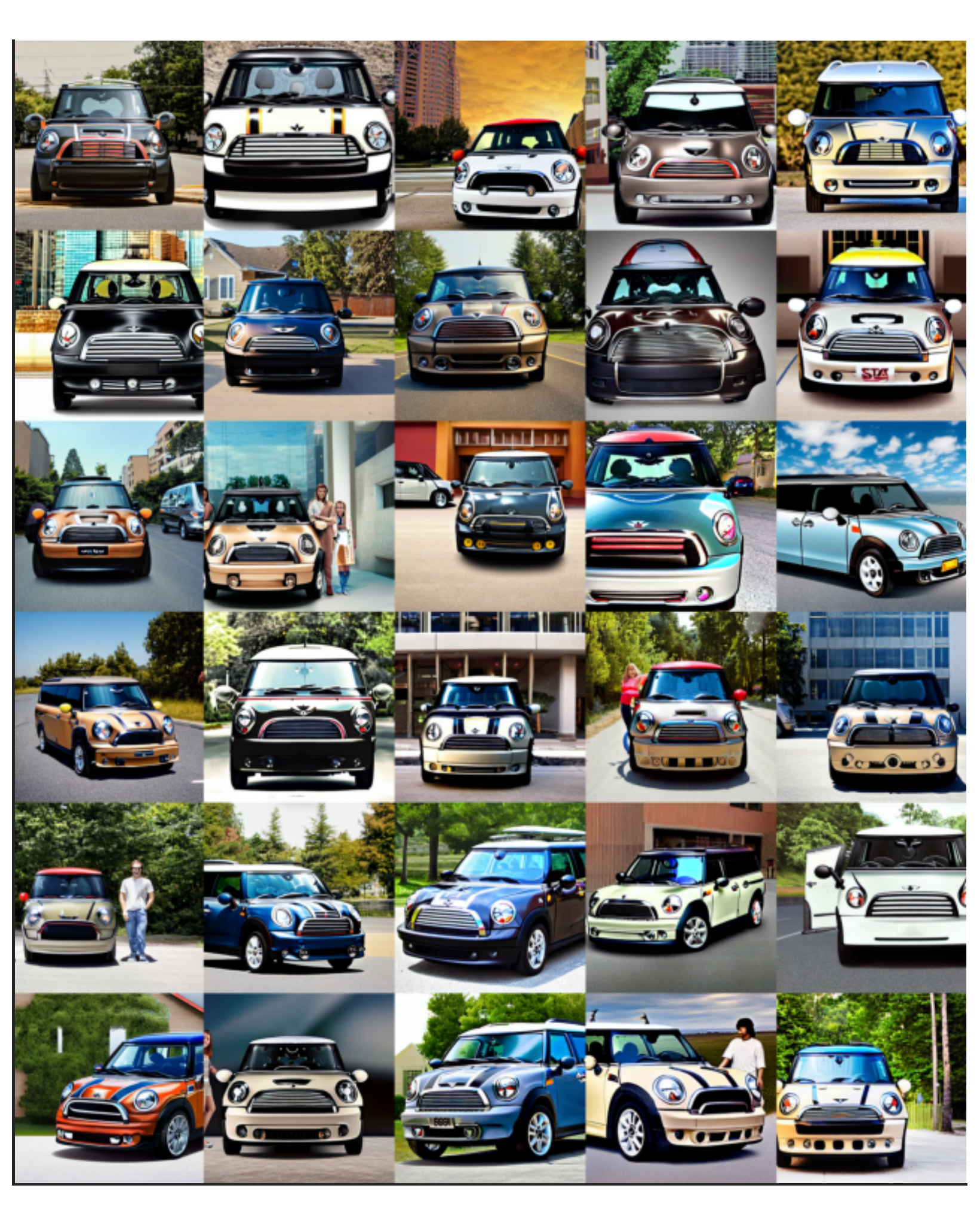}
	\end{center}
	\caption{30 samples from prompt ``front view of   (minivan:1.5) in front of sunny people.'' that are misclassified as pickups by a ViT-L/32 \cite{dosovitskiy2021an}. Please note that often, there are no ``people'' in the background, indicating a shortcoming in the text-to-image model.}
	\label{Figure:classifications_pickup}
\end{figure*}

\begin{figure*}[tb]
	\begin{center}
		\includegraphics{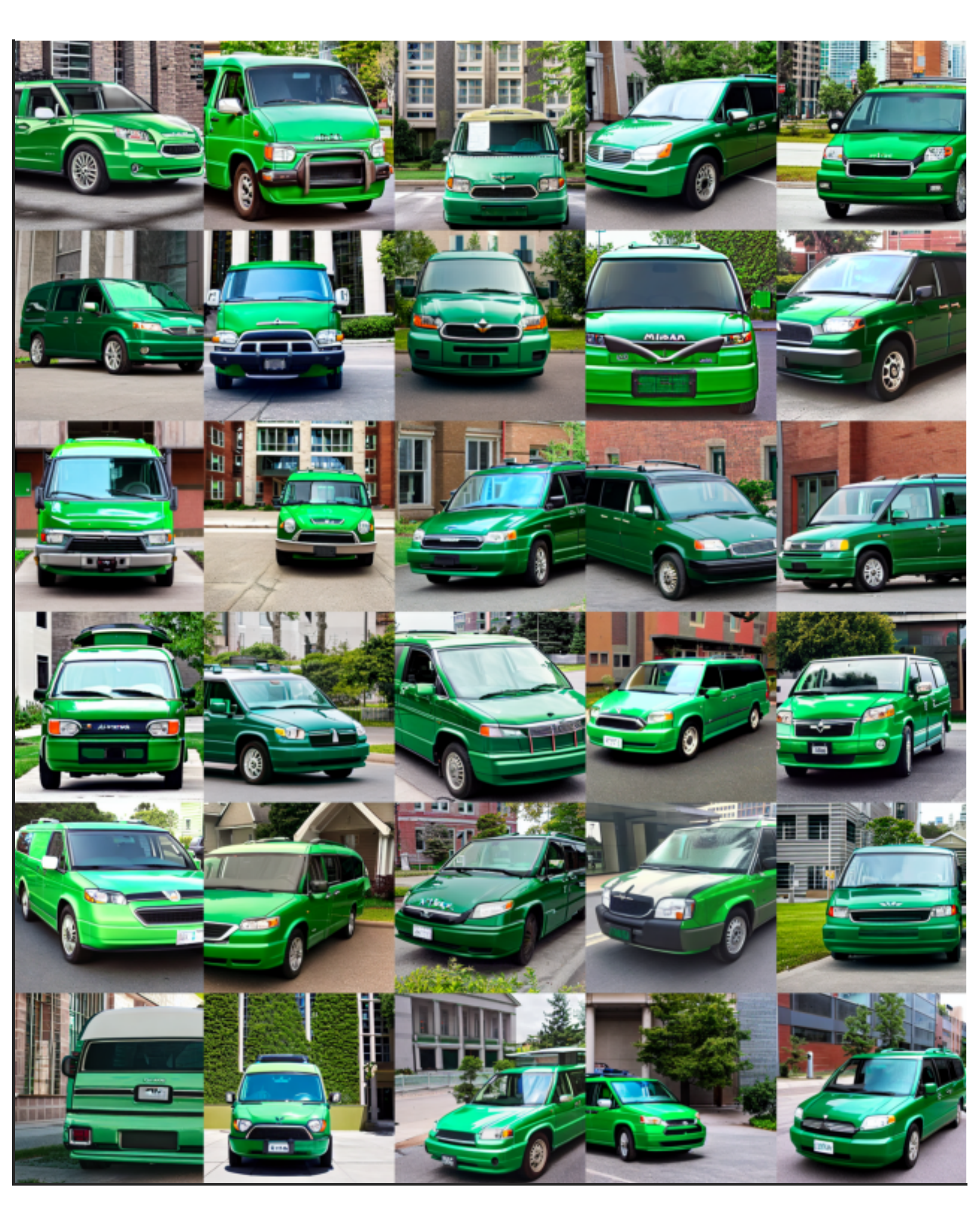}
	\end{center}
	\caption{30 samples from prompt ``front view of large green (minivan:1.5) in front of lightning house.'' that are misclassified as police-vans by a ResNet50. Please note that ``lightning'' is typically interpreted as a well illuminated scene and not as an actually lightning.}
	\label{Figure:classifications_police_van}
\end{figure*}

\begin{figure*}[tb]
	\begin{center}
		\includegraphics{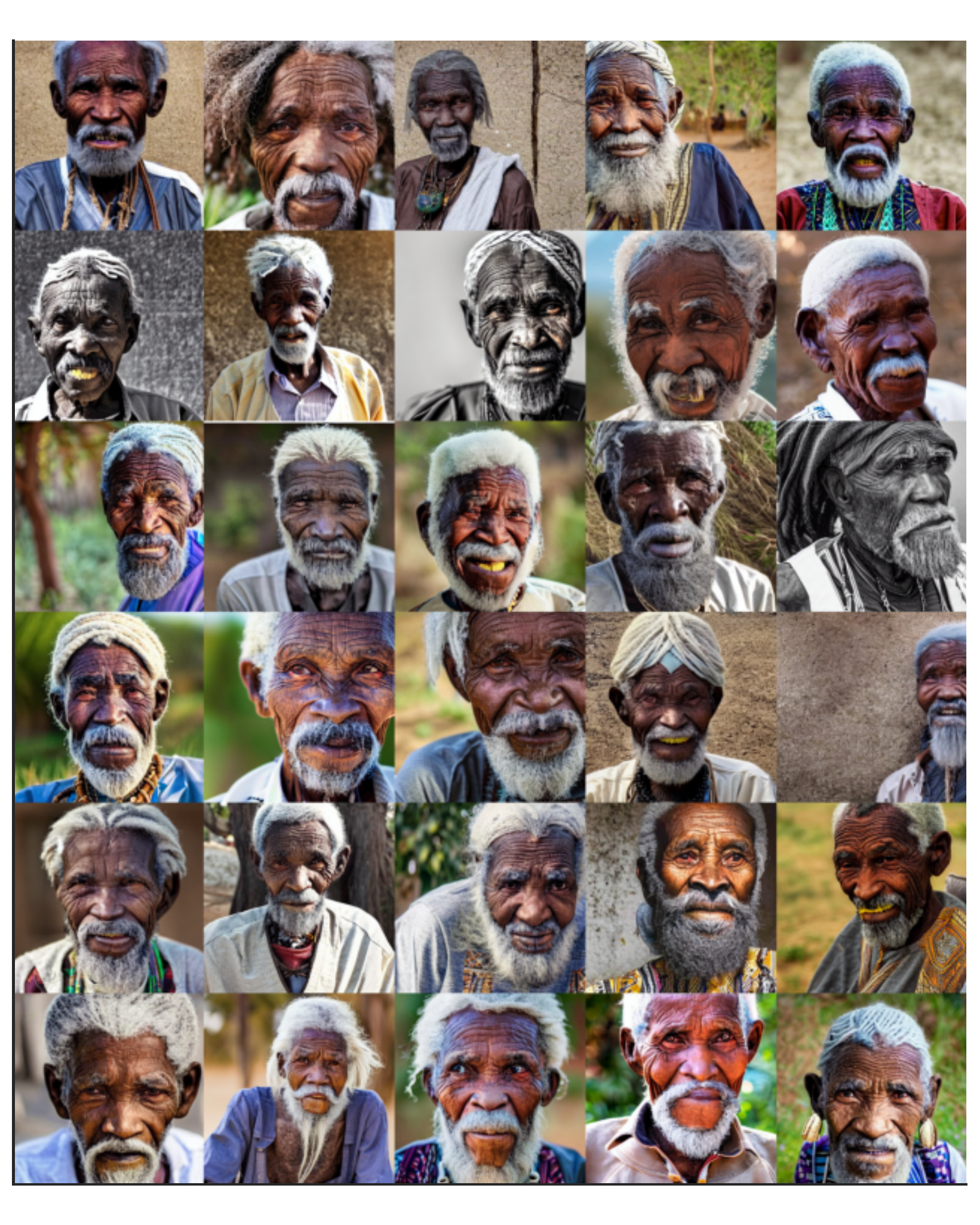}
	\end{center}
	\caption{30 samples from prompt ``A old male african (person:1.5) with long hairs in front of background'' that get a higher score for ape than for homo by a MLP-Mixer-B/16 \cite{tolstikhin2021} trained on ImageNet21k. We note that the samples from the text-to-image model are relatively similar and not fully representative of ``old male african persons with long hairs''; this systematic error thus presumably correspond to a narrower subgroup than specified by above prompt.}
	\label{Figure:classifications_ape}
\end{figure*}

We illustrate 30 samples of source class ``minivan'' misclassified as ``snowplow'' (Figure \ref{Figure:classifications_snowplow}), ``pickup'' (Figure \ref{Figure:classifications_pickup}), and ``police van'' (Figure \ref{Figure:classifications_police_van}). Moreover, we illustrate 30 samples of source class ``person'' misclassified as ``ape'' (Figure \ref{Figure:classifications_ape}).

\clearpage

\end{document}